\documentclass[conference]{IEEEtran}
\usepackage{times}

\usepackage[utf8]{inputenc}
\usepackage[T1]{fontenc}

\usepackage{amsmath}
\usepackage{amsfonts}
\usepackage{bm}

\usepackage{graphicx}
\usepackage{booktabs}

\usepackage[caption=false,font=footnotesize]{subfig}

\usepackage{algorithm}
\usepackage{algpseudocode}

\usepackage{microtype}
\usepackage{nicefrac}
\usepackage{xcolor}
\usepackage{url}

\usepackage[numbers]{natbib}

\usepackage[bookmarks=true]{hyperref}

\usepackage[capitalize,noabbrev]{cleveref}

\usepackage{listings}
\usepackage{multirow}
\usepackage{tabularx}

\newcommand{\para}[1]{\noindent\textbf{#1}}

\DeclareMathOperator{\diag}{diag}
\DeclareUnicodeCharacter{00AD}{}

\definecolor{darkyellow}{RGB}{180,150,0}

\usepackage{times}

\usepackage[utf8]{inputenc}
\usepackage[T1]{fontenc}

\usepackage{amsmath}
\usepackage{amsfonts}
\usepackage{bm}
\allowdisplaybreaks

\usepackage{booktabs}
\usepackage{multirow}
\usepackage{array}    
\usepackage{tabularx}

\usepackage{graphicx}
\usepackage[caption=false,font=footnotesize]{subfig}

\usepackage{algorithm}
\usepackage{algpseudocode}

\usepackage{microtype}
\usepackage{nicefrac}
\usepackage{xcolor}
\usepackage{url}

\usepackage{listings}
\usepackage[numbers]{natbib}

\usepackage[bookmarks=true]{hyperref}

\usepackage[capitalize,noabbrev]{cleveref}

\begin{document}

\title{QuickLAP: Quick Language-Action Preference Learning for Semi-Autonomous Agents}


\author{
\authorblockN{Jordan Abi Nader, David Lee, Nathaniel Dennler, Andreea Bobu}
\authorblockA{
\{jordancn, dlee888, dennler, abobu\}@mit.edu\\
MIT\\
United States of America
}
}


%

\maketitle

\begin{abstract}
Robots must learn from both what people do and what they say, but either modality alone is often incomplete: physical corrections are grounded but ambiguous in intent, while language expresses high-level goals but lacks physical grounding.
  We introduce \textbf{QuickLAP}: \emph{Quick Language–Action Preference learning}, a Bayesian framework that fuses physical and language feedback to infer reward functions in real time. 
  Our key insight is to treat language as a probabilistic observation over the user’s latent preferences, clarifying which reward features matter and how physical corrections should be interpreted. 
  QuickLAP uses Language Models (LMs) to extract reward feature attention masks and preference shifts from free-form utterances that are combined with physical feedback in a closed-form update rule. This enables fast, real-time, and robust reward learning that handles ambiguous feedback. In a robotic manipulation and a semi-autonomous driving simulator, QuickLAP reduces reward learning error by over 70\% compared to physical-only and heuristic multimodal baselines. User studies further validate our approach: participants found QuickLAP significantly more understandable and collaborative, and preferred its learned behavior over baselines.
  
\end{abstract}

\IEEEpeerreviewmaketitle

\section{Introduction}
\begin{figure}[t]
    \centering
    \includegraphics[width=\linewidth]{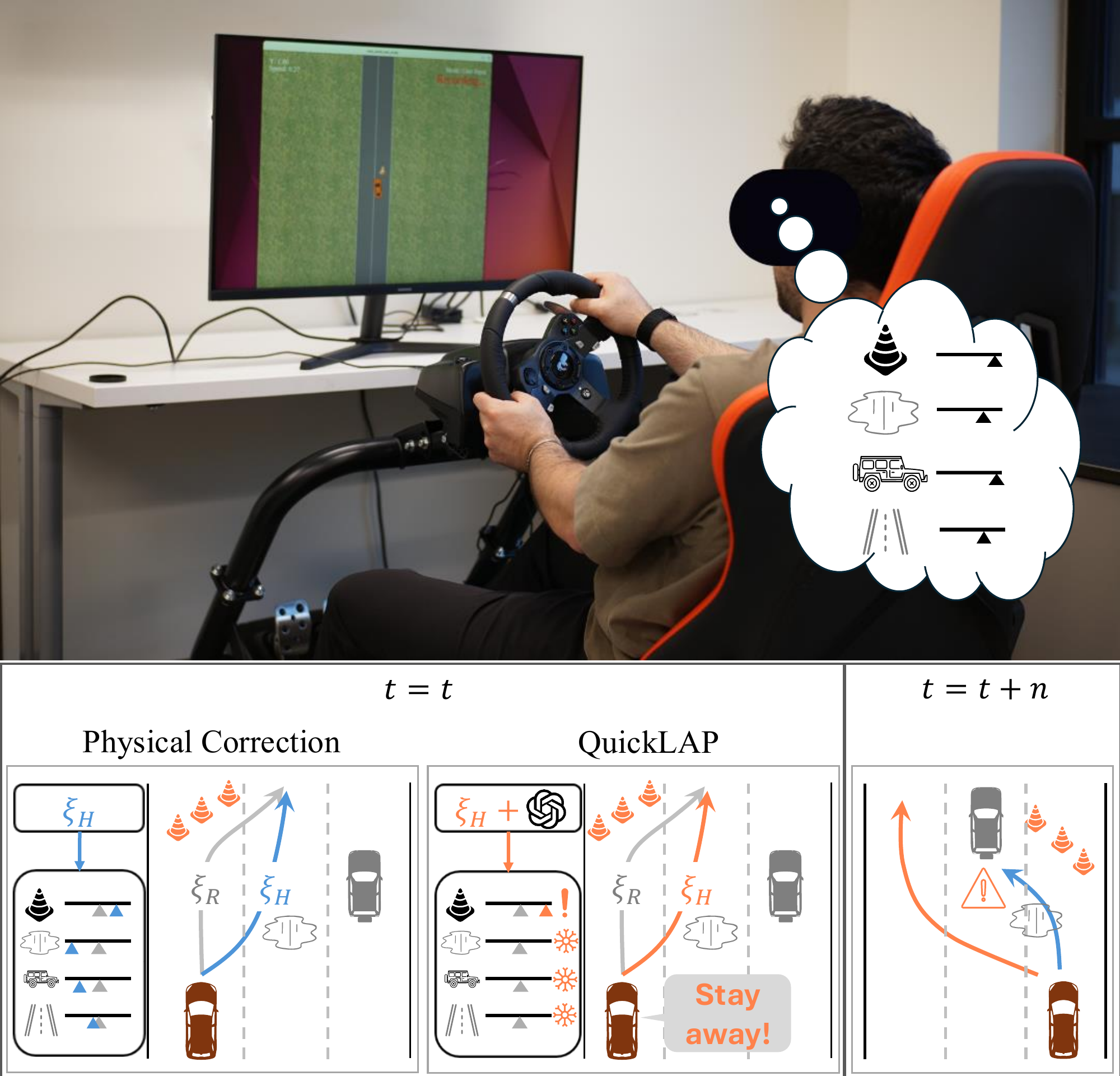}
    \caption{(Top) User Study Setup. Participants controlled the virtual car using a gaming steering wheel. (Bottom) QuickLAP fuses physical corrections and natural language for improved reward learning. (Bottom-Left) A human's physical correction (blue) leads to an update that incorrectly changes the reward on multiple correlated features. (Bottom-Middle) QuickLAP combines this physical signal with concurrent language input, using an LLM to clarify intent and produce a more accurate reward update prioritizing cone avoidance. (Bottom-Right) In future scenarios, QuickLAP generates safer trajectories (orange) than the baseline (blue). \label{fig:front_fig}}
\end{figure}

Humans reveal their preferences through both what they do and what they say. Consider the semi-autonomous vehicle approaching a construction zone in Fig.~\ref{fig:front_fig}: the user could physically nudge the steering wheel away, or say ``Stay away from the cones!''. Each of these signals is only part of the story: physical corrections offer precise, reactive adjustments but are opaque in intent, while language clarifies \textit{what} matters to the user but lacks the physical grounding needed to guide behavior on its own. To adapt effectively, robots must learn to interpret both together, in context and in real time.

Inverse Reinforcement Learning (IRL) methods that learn from physical corrections can adapt quickly and infer a reward function over task-relevant features (e.g. distance to obstacles, etc.) that reflects the user’s underlying preferences~\cite{bajcsy2017phri}. But while these corrections are grounded and precise, they are inherently ambiguous: the same steering adjustment in Fig.~\ref{fig:front_fig} could reflect a desire to avoid construction zones, to change lanes, or to stay close to puddles. Natural language can help resolve this ambiguity. An utterance like ``Stay far from the cones!'' makes the user’s intent explicit, revealing preferences that are hard to infer from motion alone. But language has its own limitations: it can be vague (``Stay away!''), underspecified, or context-dependent. On its own, it may fail to convey exactly how the robot should act. Crucially, we observe that these two feedback channels are complementary: language clarifies the intent behind a correction, and the correction grounds and disambiguates the language. Together, they offer a more complete signal than either can provide alone.

Yet few existing methods take full advantage of this complementarity. Some multimodal robot learning methods rely on large datasets of paired trajectories and language~\cite{lynch2021languageIL, lynch2022language, sharma2022correcting, stepputtis2020language, bucker2023latte}, limiting online adaptation. Others assume language inputs are precise, unambiguous, and self-contained~\cite{cui2023lang, Karamcheti2021lila}, rather than interpreting them in conjunction with behavior. As a result, these methods fall short in exactly the settings where joint feedback should be most useful: ambiguous, fast-changing, and underspecified environments.

Our key insight is that language can serve as a probabilistic observation over the user’s latent reward, not just \textit{describing} what matters, but \textit{modulating} how physical corrections should be interpreted. Language plays two roles: it identifies which features are relevant to the user’s intent, and it suggests how those preferences should shift. This perspective enables a principled fusion of modalities that resolves ambiguity and supports rapid online learning.

We introduce \textbf{QuickLAP}: \emph{Quick Language–Action Preference learning}, a closed-form Bayesian framework for online reward inference from joint physical and language feedback. For each intervention, a Language Model (LM) processes the user’s utterance in context to produce a feature attention mask (what matters), a proposed reward shift (how behavior should change), and confidence weights (how certain the LM is). 
These signals are fused with a Boltzmann-rational model of the correction and an attention-weighted prior to produce a Gaussian posterior over reward parameters. This yields an efficient MAP update rule that extends IRL to multimodal feedback, robustly handles ambiguity, and supports real-time adaptation. 
Importantly, QuickLAP is robust to ambiguity in either channel: vague utterances are interpreted in the context of physical corrections, and unclear corrections are disambiguated by language. Rather than assuming clean or isolated inputs, QuickLAP exploits the natural interplay between language and action to infer what the user wants efficiently and in real time.

To summarize, our main contributions are: 
1) an efficient Bayesian framework for jointly interpreting physical corrections and natural language feedback in real time;
2) a LM-based semantic parser procedure that maps free-form language to structured reward signals without task-specific training;
3) extensive simulation results showing large reductions in reward-inference error;
4) two user studies demonstrating improved understanding, collaboration, and user-preferred behaviors for QuickLAP over baselines.
Beyond driving and robot manipulation, QuickLAP offers a general framework for preference learning in any domain where users can show and tell, from assistive robots to collaborative drones, enabling faster, more intuitive, and more personalized human-robot interaction.

\section{Related Work}
\noindent\textbf{Learning from Physical Human Feedback.}
Early work in IRL focused on inferring reward functions from demonstrations, framing the problem as a linear program~\cite{ng2000irl}, or incorporating max-margin~\cite{abbeel2004apprentice}, maximum-entropy~\cite{maxent,bobu2018learning}, and Bayesian~\cite{Ramachandran2007birl} principles to address ambiguity over optimal behavior.
Although these methods traditionally assumed a fixed batch of demonstrations, later work recognized the value of \emph{interactive} feedback provided during task execution~\cite{bajcsy2017phri, jain2015coactive,zurek2021casa,CLEA}. Bajcsy et al. ~\cite{bajcsy2017phri} proposed learning from
\emph{physical corrections}, kinesthetic nudges that convey gradient-like information about user preferences in real time. Follow-up work extended these ideas to handle both offline and online feedback modalities through a unified framework~\cite{mehta2024unifying}, while Jain et al.~\cite{jain2015coactive} explored coactive learning from incremental physical corrections in manipulation tasks. While physical feedback offers grounded and precise input, it often affects multiple features simultaneously, making it difficult to infer the user’s true intent~\cite{bajcsy2018oneatatime, bobu2020quant,hedlund2025learning}. Our method tackles this shortcoming by exploring richer feedback modalities that can clarify user intent, such as natural language.

\smallskip
\noindent\textbf{Language in Robot Learning.}
Natural language is an expressive interface for communicating goals and preferences to robots. While early systems mapped language to symbolic actions~\cite{tellex2011understanding}, recent methods use LLMs to interpret and guide behavior~\cite{ichter2022saycan, huang2022language, stepputtis2020language}. However, most approaches assume access to large paired datasets of trajectories and utterances~\cite{lynch2021languageIL,lynch2022language,sharma2022correcting,bucker2023latte,peng2024algae,peng2024preference,sripathy2022cassie,yang2025trajectory} or rely on offline reward supervision~\cite{ouyang2022instructgpt}. Others adapt to language at test time~\cite{cui2023lang, Karamcheti2021lila}, but treat instructions as clean, self-contained inputs, ignoring physical context. These approaches typically require extensive offline training on paired datasets, and rely on pre-training with unambiguous instructions \cite{cui2023lang,yang2025trajectory}. Fusion strategies are often heuristic, with no probabilistic treatment of uncertainty or signal agreement. In contrast, we treat language as a probabilistic observation over the user’s latent reward. Our method fuses it with physical feedback in real time, yielding a closed-form Bayesian update that supports ambiguity resolution, generalization, and efficient online learning.

\smallskip
\noindent\textbf{Online Reward Learning from Human Input.}
QuickLAP builds on prior work for online reward adaptation during task execution. Interactive methods like TAMER~\cite{knox2009tamer} and COACH~\cite{macglashan2017coach} showed how scalar feedback can shape behavior incrementally. Jain et al.~\cite{jain2015coactive} extended this to richer corrections in manipulation tasks, where humans suggest incremental trajectory improvements that are immediately incorporated. Related efforts in active preference learning~\cite{biyik2019asking,sadigh2017active,dennler2024improving} focus on generating informative queries to improve reward inference, though typically in offline or batch settings. More recent approaches incorporate language into the learning loop. Some require extensive pre-collected feedback~\cite{lee2021pebble, cui2023lang}, while others use language for reward shaping~\cite{cohen2021learning,sumers2020learning} but treat it as a standalone modality, decoupled from physical input. While recent work explores LLM-driven mode switching for assistive tasks \cite{cui2023lang,tao2025lams}, it often assumes a discrete set of behaviors rather than operating in continuous control domains. QuickLAP enables online adaptation in continuous domains by interpreting physical and linguistic feedback jointly, within a unified Bayesian framework. This allows reward inference from sparse, ambiguous inputs without large datasets or perfect instructions.

\section{Problem Formulation}
\label{subsec:scenario_interaction}
We consider a robot that adapts to a human operator’s preferences and plans under deterministic dynamics $x^{t+1} = f(x^t, u^t)$, where $x$ denotes the state (e.g., position, velocity) and $u$ the control input (e.g., acceleration, steering). To align its behavior with the human preference, the robot optimizes a linear reward function \(R_\theta=\theta^\top\Phi(\xi)\), where $\Phi(\xi) = \sum_{j=0}^{T-1} \phi(x^j, u^j)$ aggregates features $\phi(x^j,u^j) \in \mathbb{R}^d$ over a trajectory $\xi=(x^0,u^0,\dots,x^T)$. These features encode relevant aspects such as proximity to obstacles, collision risk, and speed. The parameter $\theta \in \mathbb{R}^d$ indicates the relative importance of these features. We assume the human's true preference weights $\theta^\star$ are \textit{unknown} to the robot.

The robot plans using Model Predictive Control (MPC)~\cite{garcia1989MPC}. At each timestep \(t\) the robot generates a trajectory \(\xi_R^t\) by solving $\xi_R^t = \arg\max_\xi \hat{\theta}^t{}^\top \Phi(\xi)$ using its current preference estimate $\hat{\theta}^t$. A human may intervene with a physical correction $u_H^t$ (e.g., a nudge) and/or a natural-language utterance $l^t$, producing a corrected trajectory $\xi_H^t$ (e.g., via trajectory deformation~\cite{bajcsy2017phri, dragan2015deformation}. The robot's task is to infer $\theta^\star$ online from $\xi_H^t$ and $l^t$.

\smallskip
\noindent\textbf{Learning from Physical Corrections.}
When the human provides a physical correction, we follow prior work~\cite{bajcsy2017phri} and model the likelihood of observing their induced trajectory $\xi_H$ given the robot's current trajectory $\xi_R$ and hidden preference $\theta$ using a Boltzmann noisily-rational model~\cite{baker2007goal, maxent}:
\begin{equation}\label{eq:boltzmann_likelihood_problem_formulation}
\begin{split}
P(\xi_H \mid \xi_R,\theta)
&\approx \exp\!\Big(\theta^\top\big(\Phi(\xi_H)-\Phi(\xi_R)\big)
-\lambda\lVert \xi_H-\xi_R\rVert^2\Big)\,.
\end{split}
\end{equation}

Here, the term $\theta^\top(\Phi(\xi_H)-\Phi(\xi_R))$ represents the relative improvement induced by the human’s correction over the robot’s current trajectory, while the quadratic term $\lambda\|\xi_H-\xi_R\|^2$ models their reluctance to make large corrections (effort minimization), with $\lambda > 0$ controlling this trade-off.

In the physical-only setting, this likelihood is combined with a Gaussian prior to yield a MAP update that adjusts the reward parameters in the direction of the feature difference $\Delta\Phi = \Phi(\xi_H) - \Phi(\xi_R)$, following~\cite{bajcsy2017phri}.
\begin{equation} \label{eq:physical_only_update_map}
\hat{\theta}^{t+1} = \hat\theta^t + \Sigma_{prior} \Delta\Phi \enspace.
\end{equation}
This update adjusts the preference weights $\hat\theta^t$ in the direction of the observed feature change $\Delta\Phi$, scaled by the prior covariance $\Sigma_{prior}$, reflecting learning from the physical intervention.

Unfortunately, relying solely on physical feedback has three critical limitations. First, physical input is often noisy~\cite{losey2022physical}. Second, it can exhibit feature coupling, where an intended change to one feature inadvertently alters others -- for example, a correction to reduce speed might also unintentionally alter the vehicle's lateral position. Third, physical actions lack semantic expressivity: they cannot easily disambiguate the underlying intent (e.g., was a slowdown near a construction zone due to general safety, concern for workers, or adherence to a perceived speed limit?) nor convey more abstract preferences related to comfort or style. These shortcomings create an ambiguity gap between what humans intend and what robots interpret. QuickLAP addresses this gap by fusing physical corrections with natural language explanations, creating a multimodal framework that captures both the precise \textit{what} of physical corrections and the contextual \textit{why} of verbal explanations.

\section{Quick Language-Action Preference Learning}

QuickLAP is a Bayesian framework that jointly interprets physical and language input to infer user preferences $\theta$ in real time. 
Our approach extends established Bayesian treatments of physical interaction~\cite{bajcsy2017phri} to the multimodal domain through a LM-mediated language likelihood, achieving both computational tractability and enhanced expressivity.
The core principle of QuickLAP leverages the complementary nature of these feedback types: physical actions demonstrate \emph{what} the user wants through trajectory adjustments, while language clarifies \emph{why} by revealing the intent behind those corrections. 
This interpretation allows QuickLAP to achieve preference learning that is both responsive to immediate corrections and grounded in understandable user rationales. 

Formally, our goal is to compute the posterior distribution over preference parameters $P(\theta \mid \xi_H, \xi_R, l)$ given the robot's proposed trajectory $\xi_R$, the human's corrective trajectory $\xi_H$, and natural language $l$. Using Bayes' rule, this posterior becomes $P(\theta \mid \xi_H, \xi_R, l) \propto P(\xi_H, l \mid \xi_R, \theta) P(\theta)$. We factorize the joint likelihood into:
\begin{equation} \label{eq:posterior_components_intro}
P(\theta \mid \xi_H, \xi_R, l) \propto \underbrace{P(\xi_H \mid \xi_R, \theta)}_{\text{Physical Likelihood}} \cdot \underbrace{P(l \mid \xi_H, \xi_R, \theta)}_{\text{Language Likelihood}} \cdot \underbrace{P(\theta)}_{\text{Prior}}\enspace.
\end{equation}
The physical likelihood is given by the Boltzmann model in Eq.~\eqref{eq:boltzmann_likelihood_problem_formulation}. We next define the language likelihood and prior models, then derive our closed-form MAP update rule for $\theta$.

\subsection{Learning from Language Input}
To incorporate language input $l$, we interpret it as a structured signal about preference changes and define a likelihood function based on this interpretation. In our implementation, this interpretation is produced by a dual-language-model (LM) framework that is motivated by our Bayesian framework (described in Appendix A). However, the learning formulation itself is agnostic to the specific language model.

\subsubsection{Language Processing with LLMs}
We decompose language interpretation into two components:  (1) identifying which features the utterance attends to, and (2) determining how those features should change.

\smallskip
\noindent\textbf{Attention Language Model ($\mathrm{LM}_\mathrm{att}$).} We use a language model to identify which features are relevant to the human's utterance, producing an attention mask $r \in \{0,1\}^d$. Our insight is that when humans give corrections, they typically focus on specific behavioral aspects rather than addressing all simultaneously~\cite{bajcsy2018oneatatime}. For instance, ``slow down near schools'' addresses speed-related features while implicitly accepting current lane position and route choices.
Given language correction $l$ and interaction context $c = \bigl(\,\Delta\Phi,\theta^t,\textit{environment\_description})$ capturing the induced feature difference, the robot's current reward estimate, and a description of the environment, we generate $r = \mathrm{LM}_\mathrm{att}(l, c)$. Each component $r_i$ indicates whether feature $\phi_i$ is relevant to the current feedback. This attention mechanism will be crucial for modulating our prior (Sec.~\ref{subsec:conditional_prior_method}).

\smallskip
\noindent\textbf{Preference Language Model ($\mathrm{LM}_\mathrm{pref}$).}
The second LM determines how attended features should change. Given the utterance $l$, and context $c$, we generate $(m, \mu) = \mathrm{LM}_\mathrm{pref}(l, c, r)$, where $m \in [0,1]^d$ is a confidence vector indicating how precisely the language specifies an update for each feature, and $\mu \in \mathbb{R}^d$ is a shift vector representing the update's magnitude and direction.
Importantly, $m$ (and $\mu$) are distinct from $r$. The attention mask $r$ identifies \textit{which features to attend to} for a potential update, influencing the prior. The confidence $m$ and shift $\mu$ determine \textit{how much and in what direction} the language suggests $\theta$ should change for those attended features, influencing the likelihood. 

For example, when an ambiguous utterance such as “Stay away!” is applied to a correction involving both a puddle and a cone, the attention mask may mark both features as relevant ($r_{\text{car}}=r_{\text{cone}}=1$), but low specificity leads to a moderately confident $m$ and a conservative shift $\mu$. In contrast, a more specific utterance like “Stay away from cones” yields focused attention ($r_{\text{cone}}=1$) with high confidence and a strong shift, enabling a decisive update aligned with the user’s intent.

In this work, we instantiate both components using an off-the-shelf large language model (LLM). However, our framework does not depend on this choice: future work could replace or fine-tune these models while preserving the same probabilistic interface for integrating language with physical feedback.

\subsubsection{Language Likelihood}
\label{subsubsec:language_likelihood_method}
Given the structured outputs of our dual-LLM framework, we now define a likelihood model for language in Eq.~\eqref{eq:posterior_components_intro}. Modeling $P(l \mid \xi_H, \xi_R, \theta)$ directly would require specifying how humans generate free-form utterances conditioned on trajectories and hidden preferences, which is not feasible. Instead, we introduce a latent \emph{proxy variable} $\mu^t = \theta - \theta^t$, representing the reward shift that the utterance is intended to convey. 

Intuitively, the human observes the robot’s current trajectory $\xi_R$, which reflects the reward weights $\theta^t$ that the robot optimized. They compare this behavior to their own preferences $\theta$, and select an utterance $l$ that, if understood, would shift the robot’s parameters from $\theta^t$ towards $\theta$. In this sense, the utterance implies an average desired change $\mu = \theta - \theta^t$. Our $\mathrm{LM}_\mathrm{pref}$ acts as an estimator, producing $\mu^t$ and confidence $m^t$ describing this latent shift. We treat $\mu^t$ as a noisy observation of $\mu$, which gives a Gaussian likelihood:
%
\begin{equation} \label{eq:language_likelihood_overall}
P(\mu^t \mid \theta, \theta^t) = \mathcal{N}(\mu^t; \theta - \theta^t, \diag(\sigma^t_L(m^t)))\enspace,
\end{equation}
where $\sigma_L$ modulates the standard deviation as a function of the confidence scores $m_i^t$. Intuitively, if confidence is high, we trust the LM-generated shift $\mu_i^t$ is close to the true corrective shift $\theta_i - \theta_i^t$, so the variance should be small, anchoring the distribution tightly to the mean. Conversely, if confidence is low, the variance should be large, reducing the influence of language in favor of relying on the physical correction. We model this by choosing $\sigma_L: [0,1] \rightarrow \mathbb{R}^+$ such that $\sigma_L(m) \to 0$ for $m \approx 1$, and $\sigma_L(m) \to 1/\epsilon_{var}$ for $m \approx 0$. One choice for $\sigma_L^2$ is, for instance, $\sigma_L(m) = k\cdot \frac{(1-m)}{\epsilon_{var} + m}$, where $\epsilon_{var}$ is a small constant for numerical stability. 
The diagonal structure reflects the assumption that language often addresses features independently (e.g., 'slow down and stay centered' addresses speed and lane position separately), which allows us to factorize the likelihood across dimensions:
$P(\mu^t \mid \theta, \theta^t) = \prod_{i=1}^d P(\mu_i^t \mid \theta_i, \theta_i^t)$.

The likelihood in Eq.~\eqref{eq:language_likelihood_overall} provides the language component of the posterior in Eq.~\eqref{eq:posterior_components_intro}. Rather than modeling the raw utterance distribution $P(l \mid \xi_H, \xi_R, \theta)$ directly, we capture its effect through the proxy shift with tractable likelihood $P(\mu^t \mid \theta, \theta^t)$, which anchors free-form utterances in reward space and allows them to be combined with physical feedback in a unified Bayesian update.

\subsection{Conditional Prior}
\label{subsec:conditional_prior_method}

One way to do multimodal fusion could be to gate the likelihood or update rule based on language. We propose a more principled approach: incorporating the attention mask $r$ directly into the prior over $\theta$, based on the intuition that humans selectively focus on relevant features \textit{before} intervening.

Specifically, we interpret $r$ as indicating which components of $\theta$ are likely to change in the current interaction. Before receiving feedback, we hold the marginal prior $\int P(\theta\mid r)p(r)\,dr$; once we parse the utterance and infer $\hat r$, we collapse this mixture to a conditional prior $P(\theta\mid\hat r)$. 
Our formulation parallels Meta-IRL~\cite{Xu2019MetaIRL}, where a latent variable selects a task-specific Gaussian prior over reward weights before a MAP update. Similarly, our mask $\hat{r}$ modulates only the prior, avoiding double-counting in the likelihood while enabling fast, structured adaptation.
Formally, once $\mathrm{LM}_\mathrm{att}$ provides $\hat{r}^t$ from utterance $l^t$, we use a conditional Gaussian prior:
\begin{equation} \label{eq:conditional_prior_overall}
P(\theta \mid \hat{r}^t) = \prod_{i=1}^d P(\theta_i \mid \hat{r}_i^t) = \prod_{i=1}^d \mathcal{N}(\theta_i; \theta_i^t, \sigma_{prior}^2(\hat{r}_i^t)),
\end{equation}
where $\theta_i^t$ is the current estimate for that feature's weight. The prior variance $\sigma_{prior}^2(\hat{r}_i^t)$ (or its inverse, precision $\Lambda_{prior,i}(\hat{r}_i^t)$) depends on the attention mask $\hat{r}_i^t$. We define the precision as $\Lambda_{prior,i}(\hat{r}_i^t) = \frac{1}{\alpha(\hat{r}_i^t+\epsilon_{prior})}$, where $\alpha$ is a base variance scale and $\epsilon_{prior} \ll 1$ ensures minimum precision.
For low attention ($\hat{r}_i^t \to 0$), precision is high (small variance $\alpha\epsilon_{prior}$), anchoring $\theta_i$ to $\theta_i^t$. For high attention ($\hat{r}_i^t \to 1$), precision is lower (variance $\alpha(1+\epsilon_{prior})$), allowing $\theta_i$ to adapt more freely. The prior model in Eq. \eqref{eq:conditional_prior_overall} is the third term in our posterior from Eq. \eqref{eq:posterior_components_intro}.

\subsection{Full Posterior and MAP Update}
\label{subsec:full_posterior_map_update}

Before deriving the full posterior and MAP update, we summarize the key assumptions underpinning our framework for tractable inference:
(1) a locally Gaussian posterior over preferences, induced by a log-quadratic physical likelihood and Gaussian language likelihood and prior; (2) stationary user preferences within each interaction episode; and (3) a reasonably calibrated dual-LLM, such that its predicted shift $\mu^t$ is an unbiased estimate of the true preference change and its confidence scores meaningfully reflect uncertainty.

\smallskip
\noindent\textbf{Posterior Formulation and MAP Update. }
Substituting the physical likelihood (Eq.~\eqref{eq:boltzmann_likelihood_problem_formulation}), language likelihood (Eq.~\eqref{eq:language_likelihood_overall}), and conditional prior (Eq.~\eqref{eq:conditional_prior_overall}) into Eq.~\eqref{eq:posterior_components_intro}, we obtain the full posterior over $\theta$. To compute the MAP estimate, we work in the log domain. Following Bajcsy et al.~\cite{bajcsy2017phri}, we approximate the physical likelihood's contribution to the log-posterior near $\hat{\theta}^t$ using the feature difference $\Delta \Phi = \Phi(\xi_H) - \Phi(\xi_R)$, yielding a linear term $\theta^\top \Delta \Phi$. The resulting log-posterior is:

\begin{align}
\log P(\theta \mid \cdot)
&= \theta^\top \Delta \Phi
- \frac{\Lambda_{\text{prior}}(\hat r^t)}{2}(\theta - \theta^t)^2 \nonumber \\
&\quad - \frac{(\mu^t - (\theta - \theta^t))^2}
{2(\sigma_L^t(m^t))^2}
+ \text{const.}
\end{align}

with $\Lambda_{prior}$ the prior precision from Sec.~\ref{subsec:conditional_prior_method} and $\sigma_{L}^2$ the language likelihood variance from Sec.~\ref{subsubsec:language_likelihood_method}.
Setting the gradient with respect to $\theta$ to zero and solving for $\theta$, we obtain the closed-form update:
\begin{equation}
   \label{eq:quicklap-MAP}
\hat{\theta}_i^{t+1} = \hat\theta_i^t + \frac{\sigma_{L,i}^2}{\Lambda_{\text{prior,i}}\sigma_{L,i}^2+1} \Delta\Phi_i + \frac{1}{\Lambda_{\text{prior,i}}\sigma_{L,i}^2+1} \mu_i^t\enspace.
\end{equation}

This is the QuickLAP MAP update rule, applied element-wise for each feature $i$.

\smallskip
\noindent\textbf{Interpretation.}
Eq.~\eqref{eq:quicklap-MAP} exhibits a Kalman‐style fusion in which an \emph{attention–adaptive precision} term decides the credit assignment between modalities.  
We define the scalar gain
\[
\kappa_i(m_i^t, \hat{r}_i^t) = \bigl[\Lambda_{prior,i}(\hat{r}_i^t)\sigma_{L,i}^{2}(m_i^t)+1\bigr]^{-1}\enspace.
\]
The MAP update can be rewritten element-wise as:
\[
\hat{\theta}_i^{t+1} = \theta_i^{t} + \kappa_i(m_i^t, \hat{r}_i^t)\,\bigl[\,\sigma_{L,i}^{2}(m_i^t)\,\Delta\Phi_i+\mu_i^t\bigr].
\]
The behavior of this update depends on the attention gate $\hat{r}_i^t$ (via $\Lambda_{prior,i}$) and language confidence $m_i^t$ (via $\sigma_{L,i}^2$).
For features with high attention score (low prior precision $\Lambda_{prior,i}$) and high language confidence (low language variance $\sigma_{L,i}^2 \approx \epsilon_{var}$), $\kappa_i \approx 1$. The update is dominated by the language-suggested shift $\mu_i^t$, while the physical correction $\Delta\Phi_i$ is suppressed by the small $\sigma_{L,i}^2$ multiplier.
Conversely, if language confidence is low (large $\sigma_{L,i}^2$), then $\kappa_i \sigma_{L,i}^2 \approx 1/\Lambda_{prior,i}$. The update gives more weight to the physical correction $\Delta\Phi_i$, scaled by the prior variance. If attention score $\hat{r}_i^t$ is low (high prior precision $\Lambda_{prior,i}$), $\kappa_i$ becomes small, and $\theta_i$ changes little from $\theta_i^t$, primarily driven by $\Delta\Phi_i$ if $\mu_i^t$ is also small or language confidence is low. This adaptive weighting allows QuickLAP to react decisively to clear interventions while remaining robust to ambiguous signals.


\begin{figure}[t]
  \centering



  \subfloat[Two-Block Environment]{%
    \includegraphics[width=0.22\textwidth]{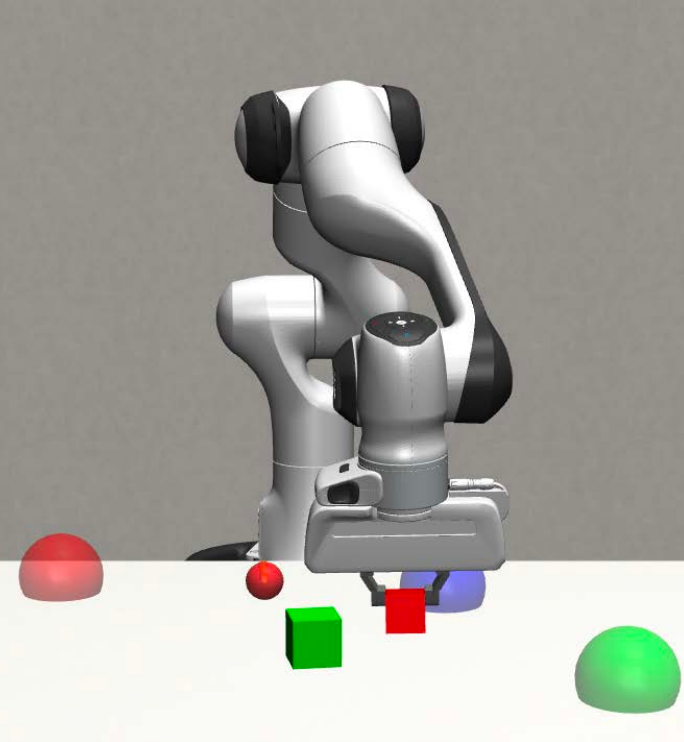}%
    \label{fig:2block}%
  }\hfill
    \subfloat[4-lane CPC (CPC-4)]{%
    \includegraphics[width=0.22\textwidth]{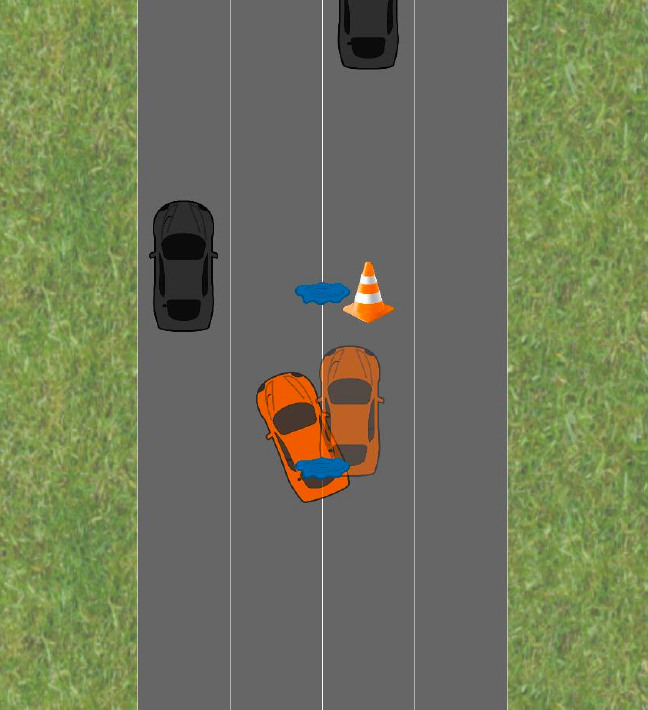}%
    \label{fig:worlds_cpc4}%
  }

  \caption{Example scenarios created from one of our four experimental driving scenarios and one robotic arm scenario.}
  \label{fig:worlds_fig}
\end{figure}

\section{Simulated Experiments}\label{sec:experiments}
We first evaluate \textbf{QuickLAP} using simulated human feedback in two domains: a robotic manipulation scenario and an autonomous vehicle scenario.

\subsection{Robot Manipulation Scenario}
\label{subsec:arm_sim}

\subsubsection{Experimental Setup}
\para{Simulator.}
We conduct our manipulation experiments in Robosuite~\cite{robosuite2020}, a physics-based simulation framework for robot learning.
We model the robot with a 6D damped second-order end-effector dynamics model.
The state is
$x = [p_x, p_y, p_z, v_x, v_y, v_z]^\top$,
where $(p_x, p_y, p_z)$ denotes the end-effector position and $(v_x, v_y, v_z)$ its velocity.
Actions
$
u = [u_x, u_y, u_z]^\top
$
correspond to desired Cartesian velocity commands.
The robot plans motions using MPC.

\smallskip
\noindent\textbf{Environment.}
The task is shown in Fig.~\ref{fig:2block}. The task requires the robot arm to transport an object from a start location (shown in red) to a goal region (shown in green) while avoiding obstacles in the \textit{Two-Block Environment}.
The green block represents an obstacle located along the nominal shortest path to the goal, while the blue region represents an area of indifference that is safe but increases path length.
This environment induces feature coupling: avoiding the obstacle often requires moving away from the goal, creating ambiguity in how physical corrections should be interpreted.

We model a simulated human with a ground-truth reward function $\theta^*$ that prioritizes obstacle avoidance over path efficiency. This simulated human provides physical corrections and language feedback at predetermined timesteps whenever the robot’s autonomous behavior conflicts with these true preferences.
Physical feedback is applied as a corrective input that perturbs the robot’s motion, often pulling the arm closer to the blue region and farther from the goal in order to avoid the obstacle.
This introduces an inherent ambiguity: a physical correction that improves safety may simultaneously degrade task efficiency.
Natural language feedback is used to clarify intent (e.g., obstacle avoidance versus goal progress), and is interpreted using the same dual-LLM pipeline and Bayesian update described earlier.

\smallskip
\noindent\textbf{Independent Variables.} We compare four approaches: 1) \textbf{Physical-Only (pHRI):} baseline that uses the physical correction to update $\hat{\theta}$ via Eq.~\eqref{eq:physical_only_update_map}, without any language input. We hypothesize that it incorrectly updates the weights for features that the human did not intend to correct.
2) \textbf{QuickLAP (Full Model):} Our proposed method, which uses both the attention mask $r$ in the conditional prior \emph{and} the language-derived shift $\mu$ and confidence $m$ in the language likelihood ~\eqref{eq:quicklap-MAP}.
We additionally compare against two ablations of QuickLAP.
3) \textbf{Masked Physical Update (Masked pHRI):} ablation of our method that incorporates the attention mask $r$ derived from language (via $\mathrm{LM}_\mathrm{att}$) into the conditional prior~\eqref{eq:conditional_prior_overall}, effectively gating which features are updated based on the physical correction $\Delta\Phi$. However, it does \emph{not} use the language-derived shift $\mu$ or confidence $m$ from $\mathrm{LM}_\mathrm{pref}$; the update relies solely on $\Delta\Phi$ for the attended features. This ablation helps identify the benefit of the attention mechanism alone.  4) \textbf{QuickLAP Language Only:} ablation of our method that isolates the effect of language input on the update function by removing the physical component from the proposed MAP update (but the LLM framework still gets access to the feature difference induced by the physical correction). This ablation is meant to test if just having the context about the person's physical input is enough for an LLM to reasonably estimate the full weight update.

During system development, we examined a ``No-Context'' baseline, where the LLM received the language utterance without any physical information. This baseline showed a drop in performance, detailed in Appendix N, because, without physical grounding, the model's semantic interpretations are not anchored to the actual feature changes occurring in the environment. Consequently, simulated utterances remained ambiguous or misaligned with the robot's state. To ensure consistent grounding, the language input in all our primary experiments is always accompanied by concurrent physical feedback. While QuickLAP is designed for this joint interpretation, we leave the exploration of entirely decoupled, language-only interventions for future work.

\smallskip
\noindent\textbf{Evaluation Metrics.}
We evaluate the accuracy of the learned preference weights $\hat{\theta}$ against the ground-truth user preferences $\theta^\star$ using \textbf{Normalized Mean Squared Error (NMSE)}, calculated as $\mathrm{NMSE}(\hat{\theta}, \theta^\star) = \frac{1}{d} \|\text{norm}(\hat{\theta}) - \text{norm}(\theta^\star)\|^2$. Lower NMSE indicates that the learned normalized preference vector is closer to the true normalized preference vector, signifying a more accurate capture of the \emph{relative importance} and \emph{direction} of the user's preferences, independent of their absolute scale.

By comparing these methods using NMSE, we aim to test two key points: (i) language provides a stronger signal than purely physical feedback for achieving reward alignment, and (ii) the combination of language and physical corrections yields the best results. In particular, the richer and more reliable interpretation of language through the shift $\mu$ and confidence $m$ contextualized by physical interventions in QuickLAP provides significant advantages beyond a simpler attention-based gating of physical updates (as in the Masked pHRI baseline), enabling more accurate recovery of rewards.

\subsubsection{Results}
Fig.~\ref{fig:sim_pilot}a reports the normalized reward-weight MSE, averaged over six utterances per method.
Physical-only learning (pHRI) and Masked pHRI shows the highest error ($0.1018 \pm 0.0001$ and $0.1000 \pm 0.0002$, respectively), indicating that physical corrections alone are insufficient to disambiguate coupled objectives such as obstacle avoidance and goal progress in this manipulation task.
In contrast, methods that incorporate language achieve substantially lower error.
Both \textbf{QuickLAP Language Only} ($0.0427 \pm 0.0043$) and \textbf{QuickLAP} ($0.0444 \pm 0.0052$) reduce reward-weight error by more than $2\times$ relative to physical-only baselines.

The strong performance of the language-only variant reflects the fact that, in this controlled setting, the language model is provided with rich physical-intervention context in its prompt, enabling it to infer the intended reward shift even without using the physical correction in the update.
The full \textbf{QuickLAP} method achieves comparable accuracy while additionally grounding the update itself in physical interaction.
This grounding does not improve accuracy in this simple domain, but is critical for robustness in settings with noisier feedback or more ambiguous language, as shown in later experiments.
Overall, these results demonstrate that language is essential for resolving ambiguity in physical corrections, while physical grounding provides a principled mechanism for maintaining reliability beyond controlled scenarios.

\subsection{Driving Scenario}
\label{subsec:driving_sim}


\subsubsection{Experimental Setup}
\label{subsec:exp}

\smallskip
\noindent\textbf{Simulator.}
In our experiments, we build on the InterACT Lab Driving Simulator~\cite{Sadigh2016PlanningFA} that uses a 4D bicycle model for vehicle dynamics. The state $x = [x, y, \theta, v]^T$ includes vehicle coordinates ($x,y$), heading ($\theta$), and speed ($v$). Actions $u = [\omega, a]^T$ are steering ($\omega$) and linear acceleration ($a$).

\smallskip
\noindent\textbf{Environments.} 
We test \textbf{QuickLAP} in four scenarios of increasing complexity (Fig.~\ref{fig:worlds_fig}). We model a simulated human with a ground truth reward function parameterized by $\theta^\star$ that highly prioritizes safety from traffic cones and other cars, is moderately concerned about avoiding puddles, and also values lane alignment (detailed in Appendix O). The simulated human would provide physical corrections and language feedback at predetermined points when the robot approached a cone too closely.

Our environments are as follows:
1) \textit{2-lane Cone (C)}: A simple two-lane road with a single cone. Physical corrections away from the cone may be confounded with lane deviation, testing whether the system can isolate the true preference (cone avoidance) from correlated features (lane alignment).
2) \textit{2-lane Cone + Puddle (CP)}: Adds a puddle to the scene. Avoiding the cone now requires entering a less-penalized puddle, making it harder to infer whether the human prefers cone avoidance over surface quality.
3) \textit{3-lane Cone + Puddle + Car (CPC-3)}: A 3-lane road with a cone, a puddle, and another vehicle. Avoiding the cone may require moving closer to the puddle or the car, testing the system's ability to disambiguate trade-offs among multiple penalized features.
4) \textit{4-lane Cone + Puddle + Cars (CPC-4)}: This world has 4 lanes and multiple cones, puddles, and several other vehicles, combining all complexities of the previous environments.

In each scenario, we assess robustness by testing QuickLAP across varied natural language phrasings with the same underlying intent to avoid traffic cones but with varying ambiguity, such as ``Be careful.'', ``Avoid the obstacle.'', ``Stay away from construction zones.'', or ``Steer clear of the cone'' (see Appendix N). Additionally, the number of human interventions were varied to observe convergence effects. We performed a total of 600 simulations to obtain the results in Fig.~\ref{fig:results}.
We ran our experiments on a single CPU and used up to 6-dimensional driving features (distance to cone, distance to puddle, lane offset, off-road risk, distance to cars, speed).

We follow the same experimental protocol, learning framework, baselines, and evaluation metrics introduced in the robotic manipulation simulation, and adapt them to a driving environment. We also evaluated a ``No-Context'' baseline (Appendix N), which confirmed that without physical grounding, language-only updates fail to anchor semantic intent to specific feature changes, especially in complex environments.

\subsubsection{Results}
\label{subsec:results_accuracy}

Our simulated driving experiments demonstrate that the trends observed in the robotic manipulation domain persist in a more complex setting, with QuickLAP consistently outperforming baseline methods in learning accurate user reward weights. The results, averaged across various language phrasings, random seeds, and distinct environmental scenarios, highlight QuickLAP's superior accuracy and convergence speed.

\smallskip
\noindent\textbf{Accuracy Across Environments.}
Figure~\ref{fig:results}a reports normalized weight MSE across four environments. QuickLAP consistently achieves lower NMSE than Masked pHRI and pHRI, with reductions of more than 50\% in moderate cases such as \texttt{CP} ($0.0761 \pm 0.1090$ vs. $0.3034 \pm 0.0236$ for Masked pHRI and $0.5941 \pm 0.0013$ for pHRI).

The QuickLAP Language Only variant closely tracks the full model (e.g., $0.2020 \pm 0.0375$ vs. $0.2058 \pm 0.0525$ in \texttt{CPC-4}), reflecting the strength of the LLM-provided signal in the update. In contrast, the full QuickLAP model combining both language and physical corrections achieves comparable accuracy while maintaining consistent performance across environments. The physical grounding provided by QuickLAP contributes to stability and prevents spurious preference shifts, which may become critical when language inputs are less reliable. We further examine this hypothesis through targeted simulation ablations that remove physical grounding from language interpretation.

\smallskip
\noindent\textbf{Convergence and Stability.}
Fig. \ref{fig:results}b illustrates the convergence behavior across methods. QuickLAP converges more rapidly than all baselines. After the first intervention, QuickLAP achieves lower NMSE than Masked pHRI or pHRI after four interventions. By the third intervention, QuickLAP stabilizes below $0.25$, while the baselines remain above $0.40$. This behavior illustrates that jointly estimating both a language-conditioned shift ($\mu$) and confidence ($m$) allows QuickLAP to incorporate human feedback efficiently while remaining grounded in physical context. In contrast, Masked pHRI provides only incremental gains over pHRI.


\section{User Evaluations}
\label{sec:study}
Our simulated results showed that QuickLAP performs well for idealized user inputs. We aim to evaluate QuickLAP's ability to generalize to realistic user inputs across different input devices. We first perform a pilot study with $N=12$ expert users in the robotic manipulation scenario (Fig.~\ref{fig:2block}) using a Connexion 3D SpaceMouse~\cite{3dconnexion2024spacemouse} to control the robot. Next, we perform a more extensive evaluation with $N=15$ non-expert users in the driving scenario (Fig.~\ref{fig:worlds_cpc4}) using a physical wheel to control the robot.
\begin{figure}
    \centering
    \includegraphics[width=1\linewidth]{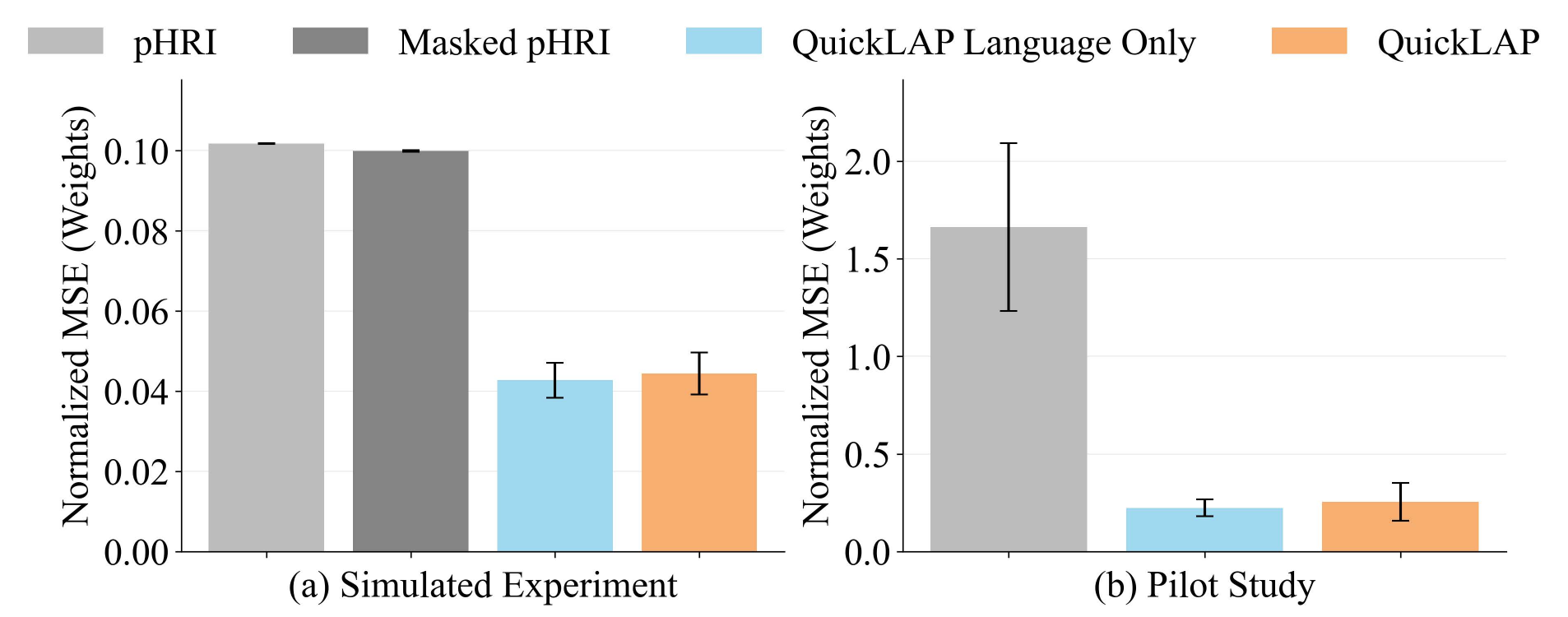}
    \caption{a) Normalized reward-weight MSE for QuickLAP and baselines in the simulated robotic manipulation scenario, averaged over 6 language inputs across 5 seeds. (b) Normalized reward-weight MSE for QuickLAP and baselines from the pilot study, averaged over 12 participant interactions. In both figures, error bars indicate the mean $\pm$ standard error of the mean (SEM) calculated across the language inputs. }
    \label{fig:sim_pilot}
\end{figure}
\subsection{Pilot Study} 

In our pilot study, each expert user interacted with a Franka robot in the block manipulation scenario. In this study, participants were first shown a desired robot trajectory that avoided the green block while moving to the green goal zone. After the participant was familiar with the optimal trajectory, the robot began executing a path, and the expert user was instructed to provide simultaneous physical and language input to correct the robot. The physical input was collected using a SpaceMouse and the user's language input was collected through a microphone and transcribed using OpenAI's Whisper model \cite{radford2022whisper}. The time for each intervention in our studies was comprised of (1) approximately one second for audio capture, (2) 1--2 seconds for speech-to-text transcription, and (3) 0.4--1.5 seconds each for 2 sequential LLM calls.
While our code was not optimized for speed this latency can be reduced to sub-second levels via parallelized `mini/nano' models and optimized, on-device speech to text transcription.

Each expert user performed three rounds of corrections, and the robot updated its estimate of the user's weights using three algorithms: pHRI, QuickLAP Language Only, and QuickLAP, described in Sec.~\ref{sec:experiments}. These three methods were presented to the user in a randomized order to control for ordering effects.

We report the NMSE of the robot's learned weights during the pilot study compared to the ground-truth weights in Fig.~\ref{fig:sim_pilot}b. We found that both variants of QuickLAP outperformed the pHRI baseline. This result indicates that the QuickLAP approach is robust to realistic physical and language input from real users.
\begin{figure}
    \centering
    \includegraphics[width=\linewidth]{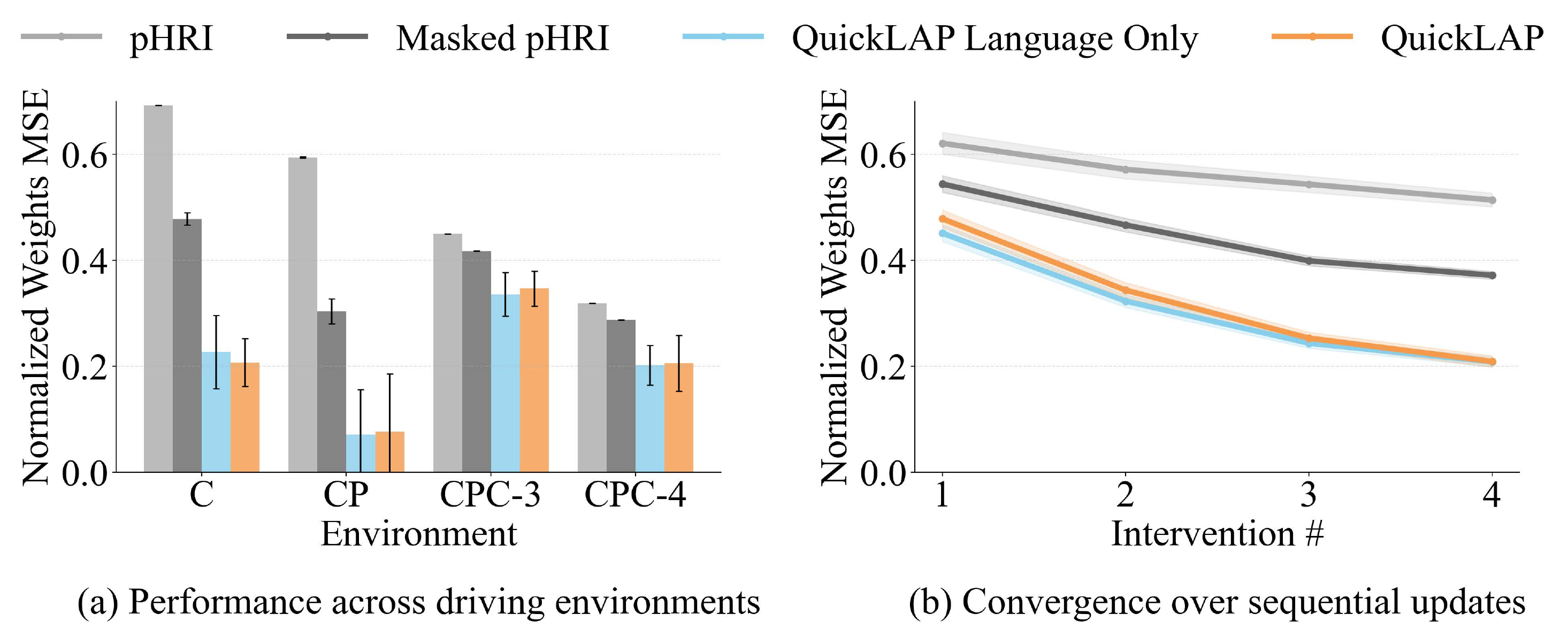}
    \caption{Comparison of adaptation methods across different environments for 4 interventions per episode. (a) Bars represent the Normalized Mean Squared Error (NMSE) for each environment, averaged over 6 language inputs. Error bars indicate the mean $\pm$ standard error of the mean (SEM) calculated across the language inputs. (b) Solid lines represent the average NMSE. Shaded regions indicate the mean $\pm$ SEM, with the SEM calculated across the 4 environments. Lower NMSE indicates better performance.}
    \label{fig:results}
\end{figure}

\subsection{Non-Expert User Study}

Following our expert user pilot study, we conducted a more extensive analysis with non-expert users to determine if QuickLAP generalizes across different user populations. We elected to use the car driving scenario for this evaluation to avoid confounding effects from the steep learning curves associated with high-dimensional robot manipulation control \cite{dhat2024using}. We performed a within-subjects user study 
where participants controlled the car using a steering wheel input as depicted in Fig.~\ref{fig:front_fig}. The experimental procedure was approved by the institutional review board.

\smallskip\noindent\textbf{Procedure.} Each participant entered a room containing the user study setup shown in Fig.~\ref{fig:front_fig}. 
Before beginning the main study, participants performed a familiarization task using the steering wheel and pedals in our setup until they were comfortable controlling the car in environment C (described in Sec. \ref{subsec:driving_sim}). Participants were assigned to a sequence of three experimental blocks. Each block sampled one environment (CP, CPC-3, CPC-4) and one algorithm (pHRI, QuickLAP Language Only, or QuickLAP). These experimental blocks were presented in a randomized and counter-balanced order.

Each block consisted of three stages. In the first stage, the participant was shown the desired driving behavior to teach the robot. In the next stage, the participant performed interventions to correct the robot using language and physical input. 
Finally, the participant observed the car performing the learned behavior.
After completing the three experimental blocks, participants filled out a questionnaire and ranked their overall preferences among the algorithms.



\smallskip
\noindent\textbf{Evaluation Metrics.} Our non-expert user study evaluated our algorithm in three ways. First, we adapted scales validated in previous work \cite{bajcsy2017phri} to assess four self-reported constructs that reflect how well a system learns from user feedback on a 7-point scale: (1) \textit{understandability}, (2) \textit{ease of use}, (3) \textit{predictability}, and (4) \textit{collaborativity}. 
Second, participants additionally ranked each algorithm relative to the other algorithms according to their overall preference. Finally, we measured performance of the vehicle behavior learned from the participant using the NMSE as in our simulated experiments.

%

\begin{figure*}[t!]
  \centering

  \subfloat[Self-reported constructs]{%
    \includegraphics[width=0.60\textwidth]{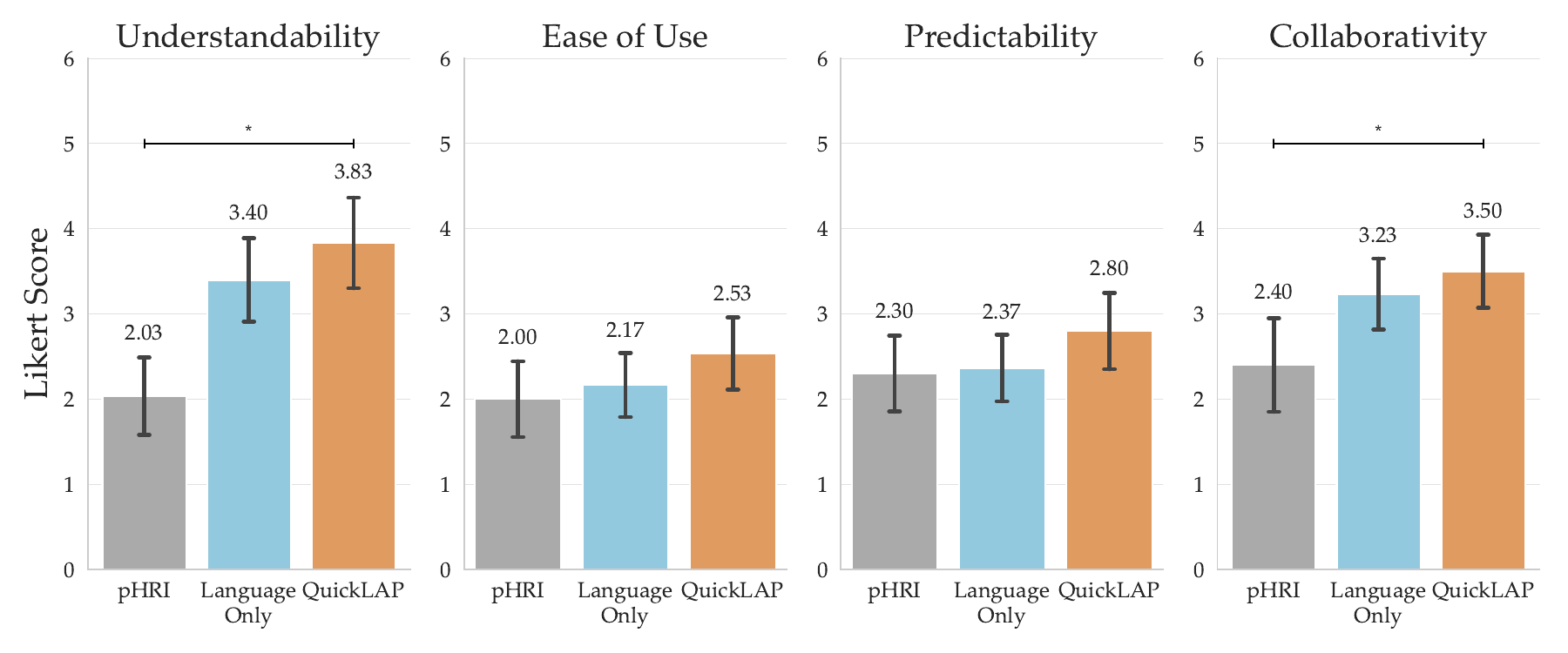}%
    \label{fig:user_study_constructs}%
  }\hfill
  \subfloat[Algorithm Ranking]{%
    \includegraphics[width=0.16\textwidth]{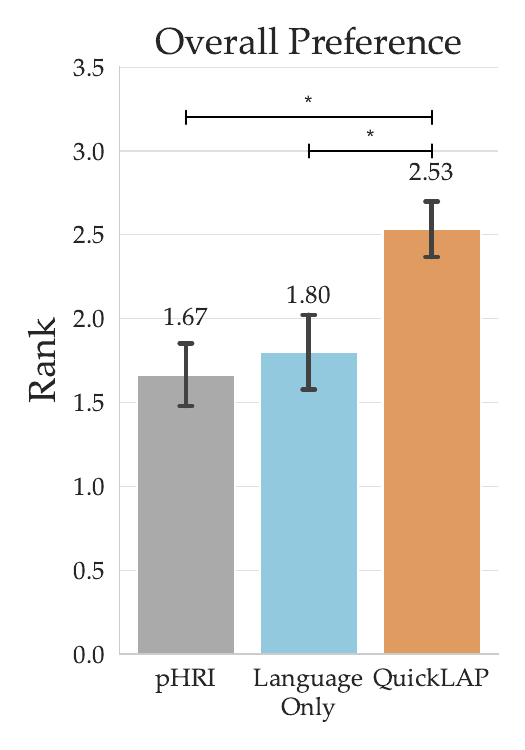}%
    \label{fig:user_study_ranking}%
  }\hfill
  \subfloat[Objective Measure]{%
    \includegraphics[width=0.16\textwidth]{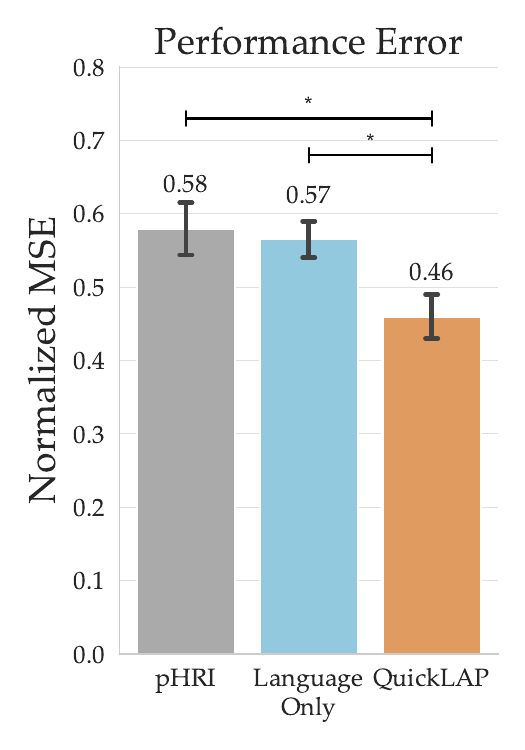}%
    \label{fig:user_study_mse}%
  }


  \caption{User study results. All error bars represent standard error. (a) Average ratings for \textit{Understandability}, \textit{Ease of Use}, \textit{Predictability}, and \textit{Collaborativity}. Higher values are better. (b) Average ranking of each algorithm. Three corresponds to the most preferred algorithm, and one corresponds to the least preferred algorithm. (c) Normalized MSE between the learned vehicle behavior and the optimal vehicle behavior. Lower values are better.}
  \label{fig:user_study_results}
\end{figure*}

\noindent\textbf{Hypotheses.} Based on the prior literature that highlights the importance of learning from multimodal signals \cite{jenamani2025feast,nanavati2025lessons,fitzgerald2022inquire,hagenow2025realm,tanjim2025help, jensen2024automated} and our simulated results, we developed the following hypotheses about our measures:

\begin{itemize}
    \item[] \textbf{(H1)} QuickLAP is (a) \textit{more understandable}, (b) \textit{easier to use}, (c) \textit{more predictable}, and (d) \textit{more collaborative} than pHRI.
    \item[] \textbf{(H2)} Participants will \textit{prefer} QuickLAP to pHRI.
    \item[] \textbf{(H3)} QuickLAP will achieve \textit{lower error} than pHRI.
\end{itemize}








\subsection{Non-Expert User Study Results}
\label{sec:user-study-results}
\noindent\textbf{H1: User-reported Constructs.} 
We performed a repeated measures ANOVA for each construct to determine significant effects. Fig.~\ref{fig:user_study_constructs} illustrates an overview of our results. We found that QuickLAP was significantly more \textit{understandable} than pHRI, $p=.023$, providing support for \textbf{H1a}. We found no significant differences for \textit{Ease of Use} (H1b) or \textit{Predictability} (H1c), but found an empirical trend that users rated QuickLAP as the easiest to use and the most predictable on average. QuickLAP was significantly more collaborative than pHRI, $p=.029$, providing support for \textbf{H1d}. Overall, these results partially support \textbf{H1}, illustrating that learning from both language and physical input results in more intuitive updates on robot behavior than learning from physical input alone, though providing two forms of input simultaneously was not necessarily easier to do.


\smallskip
\noindent\textbf{H2: Overall Preference.} Users ranked the three algorithms relative to each other at the end of the experiment. Because these ranking data are ordinal rather than continuous, we used a non-parametric Friedman test with the algorithm as a within-subjects factor and rank as the dependent variable. The average ranks of each algorithm are shown in Fig.~\ref{fig:user_study_ranking}. 
Participants ranked QuickLAP significantly higher than both Language Only, $p=.048$, and pHRI, $p=.022$. This result supports \textbf{H2}, highlighting QuickLAP's ability to incorporate language input to provide more accurate preference updates. 

\smallskip
\noindent\textbf{H3: Performance Error.} We assessed performance error by comparing the behavior the vehicle learned to the optimal behavior participants attempted to emulate, shown in Fig.~\ref{fig:user_study_mse}. 
We found that QuickLAP achieved significantly lower NMSE than both Language Only, $p=.015$, and pHRI, $p=.040$. 
This result supports \textbf{H3}, demonstrating that participants were able to teach the robot behaviors that were functionally better by using QuickLAP compared to the other algorithms.


\section{Discussion}
Our results highlight key differences between \textit{simulated} and \textit{real} human feedback. In simulation, QuickLAP Language Only performs nearly as well as the full model, because our simulated utterances correspond directly to feature-level feedback (e.g., ``avoid the obstacle'' for the cone feature). We also saw that expert users showed similar trends, where NMSE was similar for both QuickLAP Language Only and QuickLAP. In the non-expert user study, however, participants often used higher-level or indirect phrasing, such as ``move to the right lane'', which does not map directly onto the robot’s features. When this happens, our probabilistic model places more emphasis on learning from physical corrections than language input, resulting in better robustness to noisy behaviors.
The user study also elicited realistic edge cases that occur when learning from multimodal feedback, such as contradictory inputs and imprecise language. For example, one user verbally commanded the robot to ``stay away'' from an obstacle while erroneously moving the robot toward the obstacle. QuickLAP’s confidence-weighted formulation (Eq. \ref{eq:quicklap-MAP}) balanced correction and language updates to prevent the incorrect reward shifts that typically occur with physical-only updates. 
Another participant mispronounced `cone' as `coin'. Because the LLM likelihood is conditioned on physical context, where a cone is present but a coin is not, QuickLAP successfully resolved the user’s true intent.
Overall, the language-only variant’s strong simulated performance suggests that, when language is accurate, a few physical examples provided as in-context prompts could approximate the role of real corrections, potentially enabling purely voice-based preference updates. However, our user study indicates that real human feedback is often inconsistent or underspecified, underscoring the need for physical grounding in practice.

\smallskip
\noindent\textbf{Limitations and Future Work.} Despite QuickLAP's demonstrated benefits, several limitations suggest avenues for future exploration. First, we assumed that the outputs of our dual-LLM framework were reasonably calibrated, i.e., that the estimated shift $\mu^t$ and confidence $m^t$ were unbiased reflections of user intent.
While empirical results suggest this assumption is appropriate in practice, explicit calibration of LLMs remains an open challenge. Formally evaluating perturbed confidence values would determine the extent to which this assumption holds in general domains. Future work could explore in-context learning with labeled examples~\cite{ren2023robots} or conformal prediction~\cite{lindemann2023safe,zhao2024conformalized} to systematically align model uncertainty with user intent. 

Second, QuickLAP currently focuses on fusing physical corrections and language. 
Robots, however, can learn from many other human cues: gaze~\cite{admoni2017social,liang2024visarl}, facial expression~\cite{cui2021empathic,dennler2021personalizing}, gesture~\cite{brown2023gestures}, and more~\cite{stiber2022modeling, candon2024react}.  
Extending QuickLAP to incorporate additional user input modalities would enable richer, more intuitive interaction. In addition, QuickLAP focused on pre-defined features that are updated via multi-modal feedback. In many HRI settings, these features may be hard to specify, however future work may incorporate methods like language-guided abstraction~\cite{peng2024plga} or pretrained segmentation models~\cite{ravi2025sam} to dynamically generate features.

Third, in our non-expert user study, participants did not rate QuickLAP as significantly easier to use or more predictable than baselines. We believe that using the same interaction to provide feedback for all methods led to this perception. While this was important for controlling the duration of feedback for each algorithm, future work may investigate the impact of the interface design for initiating corrections with respect to their ease of use and predictability. Additionally, formal quantitative analysis on the robustness to contradictory inputs remains a limitation of this study.
Furthermore, while current LLM API latencies were sufficient for the interaction frequencies observed in our studies, future deployments on high-speed platforms could utilize local, distilled models to further minimize the delay between a user's utterance and the reward update.

 Finally, our user evaluations assumed a single optimal behavior for users to emulate. While this is realistic for driving, where people often follow the same road rules, other HRI settings like social navigation~\cite{mavrogiannis2023core}, assistive teleoperation~\cite{collier2025sense,tao2025lams}, or manufacturing~\cite{nemlekar2023transfer}, involve diverse, subjective goals. Our experiments showed QuickLAP's ability to generalize across simple manipulation tasks and driving environments, but future work may further evaluate its applicability in different domains. 
 
\smallskip
\noindent\textbf{Conclusion.} In this paper, we presented QuickLAP, a Bayesian framework that fuses physical corrections and natural language to infer human reward functions in real time. Across simulation and a user study, QuickLAP consistently improved reward learning compared to physical-only or heuristic multimodal baselines. 
Participants found our approach more understandable and collaborative, and its learned behaviors were closer to their intended goals. Together, these findings highlight the promise of learning from multiple feedback types, grounding abstract utterances directly in reward space.






%


\section*{Acknowledgments}

This research was supported in part by the MIT Energy Initiative
(MITEI) Seed Award.
The authors would like to thank the participants of our user studies for their time and valuable feedback.

\bibliographystyle{plainnat}
\bibliography{references}
\newpage
\appendix

\section{Derivations and Algorithmic Details}
\label{app:derivation}

\subsection{QuickLAP MAP Update Derivation}
\label{app:quicklap-map-derivation}

We base our derivation on the graphical model shown below.

\begin{figure}[ht]
    \centering
    \includegraphics[width=\linewidth]{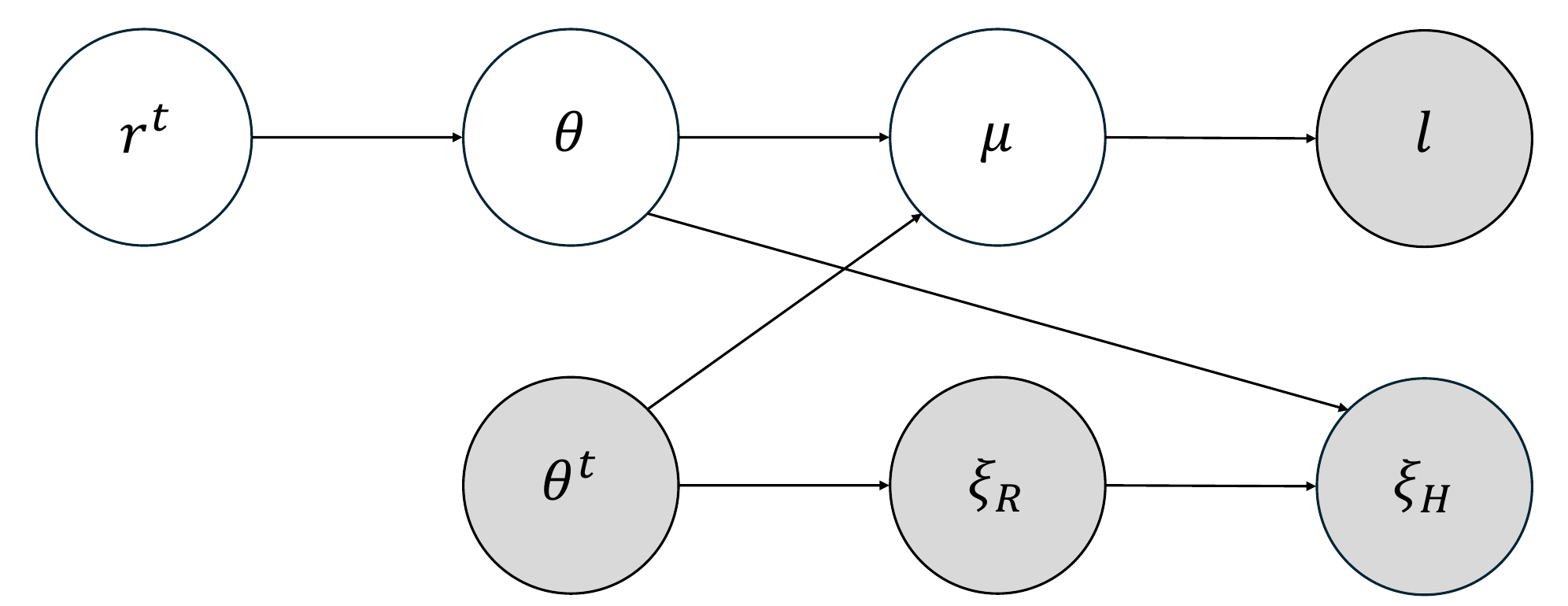}
    \caption{Graphical model for QuickLAP. The robot optimizes with current reward parameters $\theta^t$, which generate a candidate trajectory $\xi_R$. The human has latent true reward parameters $\theta$. An attention mask $r^t$ (at time $t$) selects the feature subspace currently attended, introducing a latent shift $\mu$ between human and robot rewards. Given $\xi_R$, the human may provide a physical intervention $\xi_H$ under $\theta$, and a language utterance $l$ that directly informs $\mu$. Shaded nodes are observed ${\theta^t, \xi_R, \xi_H, l}$; unshaded nodes are latent ${\theta, r, \mu}$.}
    \label{fig:placeholder}
\end{figure}

Starting from the posterior in Eq.~(3) of the main text:
\begin{equation}
P(\theta \mid \xi_H, \xi_R, l) \propto P(\xi_H \mid \xi_R, \theta) \cdot P(l \mid \xi_H, \xi_R, \theta) \cdot P(\theta \mid \hat{r})
\end{equation}
To deal with the infeasible $P(l \mid \xi_H, \xi_R, \theta)$, we introduce a latent shift $\mu$ that is informed by the language $l$ and conditioned on $\theta$ and $\theta^t$. We get:
\begin{equation}
P(\theta \mid \xi_H, \xi_R, l) \propto P(\xi_H \mid \xi_R, \theta) \cdot P(\mu \mid \theta, \theta^t) \cdot P(l \mid \mu) \cdot P(\theta \mid \hat{r})
\end{equation}
Since $l$ is independent of $\theta$ given $\mu$, the language term reduces to a factor over $\mu$; the prior $P(\mu \mid \theta, \theta^t)$ remains. This reflects the assumption that language communicates relative preference shifts rather than absolute reward parameters.

\para{Step 1: Log-Posterior Construction.}
Taking the logarithm and substituting our component models:
\begin{equation}
\begin{aligned}
\log P(\theta \mid \cdot)
&= \theta^\top \Delta\Phi - \lambda \|\xi_H - \xi_R\|^2 \\
&\quad - \frac{1}{2}(\theta - \theta^t)^T
\Lambda_{\text{prior}}(\hat{r}^t)(\theta - \theta^t) \\
&\quad - \frac{1}{2}(\mu^t - (\theta^t - \theta))^T
\Sigma_L^{-1}(m^t)(\mu^t - (\theta^t - \theta)) \\
&\quad + \text{const}
\end{aligned}
\end{equation}

where:
\begin{itemize}
\item $\Delta\Phi = \Phi(\xi_H) - \Phi(\xi_R)$ is the feature difference
\item $\Lambda_{\text{prior}}(\hat{r}^t) = \text{diag}(1/(\alpha(\hat{r}_i^t + \epsilon_{\text{prior}})))$ is attention-modulated precision
\item $\Sigma_L^{-1}(m^t) = \text{diag}(1/\sigma_L^2(m_i^t))$ is confidence-modulated precision
\end{itemize}

\para{Step 2: MAP Optimization.}
Taking the gradient with respect to $\theta$ and setting to zero:
\begin{equation}
\begin{aligned}
\nabla_\theta \log P(\theta \mid \cdot)
&= \Delta\Phi
 - \Lambda_{\text{prior}}(\hat{r}^t)\,(\theta - \theta^t) \\
&\quad - \Sigma_L^{-1}(m^t)\,\bigl(-\mu^t + (\theta - \theta^t)\bigr)
= 0
\end{aligned}
\end{equation}

\para{Step 3: Element-wise Solution.}
For each feature $i$, rearranging:
\begin{align}
\Delta\Phi_i &= \Lambda_{\text{prior},i}(\theta_i - \theta_i^t) + \sigma_{L,i}^{-2}(\theta_i - (\theta_i^t + \mu_i^t)) \\
\Delta\Phi_i &= (\Lambda_{\text{prior},i} + \sigma_{L,i}^{-2})\theta_i - \Lambda_{\text{prior},i}\theta_i^t - \sigma_{L,i}^{-2}(\theta_i^t + \mu_i^t)
\end{align}

Solving for $\theta_i$:
\begin{equation}
\theta_i = \frac{\Delta\Phi_i + \Lambda_{\text{prior},i}\theta_i^t + \sigma_{L,i}^{-2}(\theta_i^t + \mu_i^t)}{\Lambda_{\text{prior},i} + \sigma_{L,i}^{-2}}
\end{equation}

Simplifying to the final update rule:
\begin{equation}
\label{eq:map-update}
\boxed{\hat{\theta}_{i}^{t+1} = \hat{\theta}_i^t + \frac{\sigma_{L,i}^2 \Delta\Phi_i + \mu_i^t}{\Lambda_{\text{prior},i} \sigma_{L,i}^2 + 1}}
\end{equation}

\para{Attention-Dependent Prior Precision:}
\begin{equation}
\Lambda_{\text{prior},i}(\hat{r}_i^t) = \frac{1}{\alpha(\hat{r}_i^t + \epsilon_{\text{prior}})}
\end{equation}
where $\alpha$ is the base variance scale and $\epsilon_{\text{prior}} \ll 1$ ensures minimum precision.

\para{Confidence-Dependent Language Variance:}
\begin{equation}
\sigma_L^2(m_i^t) = k^2 \cdot \frac{(1 - m_i^t)^2}{(\epsilon_{\text{var}} + m_i^t)^2}
\end{equation}

\subsection{Derivation and Interpretation of the QuickLAP Update}
\label{app:quicklap_interp}

\subsection{QuickLAP Algorithm}
\label{app:quicklap-algorithm}

\begin{algorithm}[H]
\caption{QuickLAP Online Learning Update \label{alg:quicklap}}
\begin{algorithmic}[1]
\Require Human correction $\xi_H$, robot plan $\xi_R$, language utterance $l$, current weights $\hat{\theta}^t$
\Ensure Updated weights $\hat{\theta}^{t+1}$

    \State Compute physical feature difference: $\Delta\Phi \leftarrow \Phi(\xi_H) - \Phi(\xi_R)$
    \State Extract attention mask: $\hat{r} \leftarrow \text{LM}_{\text{att}}\bigl(l, \Delta\Phi\bigr)$
    \State Extract preference shift and confidence: $(\mu, m) \leftarrow \text{LM}_{\text{pref}}\bigl(l, \Delta\Phi, \hat{r}\bigr)$
    \For{$i \gets 1$ \textbf{to} $d$} \Comment{$d$ is the number of features}
        \State $\Lambda_{\text{prior},i} \leftarrow \dfrac{1}{\alpha\, (\hat{r}_i + \epsilon_{\text{prior}})}$ \Comment{attention-weighted prior precision}
        \State $\sigma_{L,i}^2 \leftarrow k^2\, \dfrac{(1 - m_i)^2}{\bigl(\epsilon_{\text{var}} + m_i\bigr)^2}$ \Comment{confidence-dependent variance}
        \State $\displaystyle \hat{\theta}_i^{t+1} \leftarrow \hat{\theta}_i^{t} + \frac{\sigma_{L,i}^2 \, \Delta\Phi_i + \mu_i}{\Lambda_{\text{prior},i}\, \sigma_{L,i}^2 + 1}$ \Comment{MAP update (Eq.~\eqref{eq:map-update})}
    \EndFor
    \State \Return $\hat{\theta}^{t+1}$
\end{algorithmic}
\end{algorithm}

\subsection{Physical--Language Trade-off Visualization}
\label{app:tradeoff}

\begin{figure}[H]
\centering
\includegraphics[width=\linewidth]{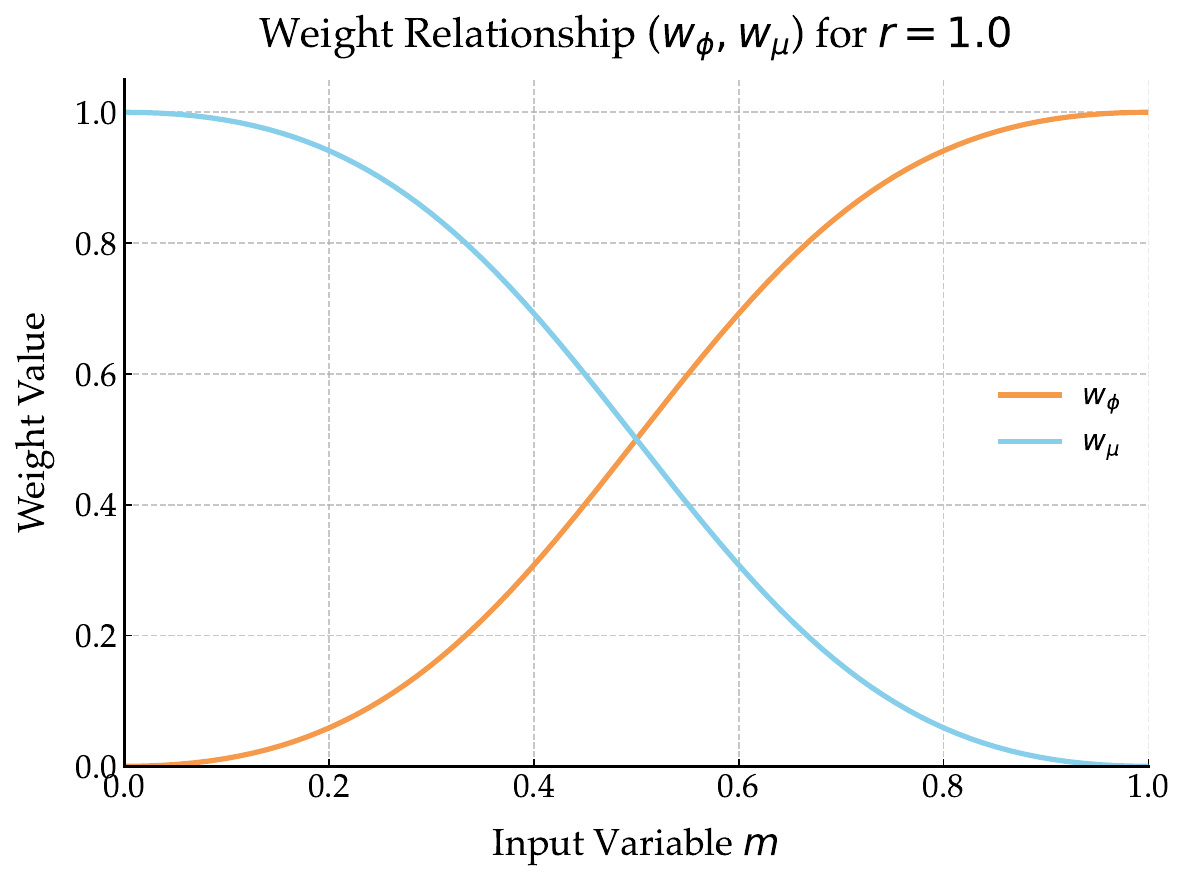}
\caption{Trade-off between physical correction weight ($w_\phi$) and language guidance weight ($w_\mu$) as a function of language confidence $m$ for high attention ($r = 1.0$). When language confidence is low ($m \approx 0$), the system relies heavily on physical corrections. As confidence increases ($m \to 1$), QuickLAP shifts toward trusting the language-suggested preference changes. This smooth transition enables robust multimodal fusion.}
\label{fig:tradeoff}
\end{figure}

Figure~\ref{fig:tradeoff} illustrates the key insight of QuickLAP: as language confidence $m$ increases, the system transitions from relying primarily on physical corrections ($w_\phi$) to trusting language-suggested changes ($w_\mu$). This adaptive weighting allows QuickLAP to handle both ambiguous and precise language feedback appropriately.

The mathematical relationship shown in the figure follows from our update rule:
\begin{align}
w_\phi &= \frac{\sigma_{L}^2}{\Lambda_{\text{prior}} \sigma_{L}^2 + 1} \\
w_\mu &= \frac{1}{\Lambda_{\text{prior}} \sigma_{L}^2 + 1}
\end{align}

where $\sigma_L^2(m) = k^2 \cdot \frac{(1-m)^2}{(\epsilon_{\text{var}} + m)^2}$ creates the smooth transition from physical-dominant to language-dominant updates. In this figure, $k = 1.0$ and $\Lambda_{\text{prior}} = 1.0$ (high attention, $r = 1.0$). The parameter $k$ controls the steepness of the transition: larger values of $k$ create sharper transitions around $m = 0.5$, while smaller values result in more gradual weight shifts across the confidence spectrum.

\subsection{Prompts, Experimental Details, and User Study Materials}
\label{app:prompts}

\subsection{LLM Prompt Templates}
\label{app:prompts-templates}
In this section of the appendix, we provide the full prompts that we used to query the LLMs in our studies.

\subsection{Attention Language Model ($\text{LM}_{\text{att}}$)}
\label{app:lm-att}

\textbf{System Message:}
\begin{lstlisting}
You are an expert in autonomous vehicle control analyzing driver interventions. In this task, a human driver has intervened to correct the behavior of a robot car and has provided an explanation of the intervention. Your task is to determine which features are relevant to a given intervention explanation, given the change in feature values of the human trajectory compared to the robot trajectory. Positive values mean that the human increased the feature value.

Note that a feature may be irrelevant even if it has a large change in value. Only output features that are relevant to the intervention explanation.

Output STRICT JSON with the single key 'gate': a list of attention gates scores (one per feature, 0.0 or 1.0). NO other keys.
\end{lstlisting}

\textbf{User Message Template:}
\begin{lstlisting}
Human Driver Intervention Explanation:
{explanation}

Current Feature Values:
- speed_desirability (How close the car's speed is to the desired speed limit): feature change after intervention: {speed_change:+.3f}, the human {speed_direction} this feature
- lane_alignment (How well the car stays centered in its lane): feature change after intervention: {lane_change:+.3f}, the human {lane_direction} this feature
- off_road (Penalty for driving off the road): feature change after intervention: {offroad_change:+.3f}, the human {offroad_direction} this feature
- cone_distance (Safe distance from traffic cones): feature change after intervention: {cone_change:+.3f}, the human {cone_direction} this feature
- car_distance (Safe distance from other vehicles): feature change after intervention: {car_change:+.3f}, the human {car_direction} this feature
- puddle_distance (Safe distance from puddles): feature change after intervention: {puddle_change:+.3f}, the human {puddle_direction} this feature

For absolutely EVERY feature above, determine:
1. How relevant is this feature to the intervention? (gate score 0.0 or 1.0)
\end{lstlisting}

\subsection{Preference Language Model ($\text{LM}_{\text{pref}}$)}
\label{app:lm-pref}

\textbf{System Message:}
\begin{lstlisting}
You are an expert in autonomous vehicle control analyzing driver interventions. In this task, a human driver has intervened to correct the behavior of a robot car and has provided an explanation of the intervention. Your reward function is the sum of the features. You want to maximize the reward function.

Feature values are between 0 and 1. Look at the feature descriptions to understand the scale of the features. Your task is to determine for EACH feature how much in magnitude should the weight of the feature be changed to support the intervention and human preference (mu between 0 and 6), and how confident you are in your decision (confidence between 0 and 1, be conservative), given the change in feature values of the human trajectory compared to the robot trajectory. Positive values mean that the human increased the feature value.

FOR EVERY FEATURE, return ONLY the values in this exact format.

OUTPUT (strict JSON, single line):
{
    'mu': [u1, u2, ... , uN],
    'confidence': [c1, c2, ... , cN]
}
\end{lstlisting}

\textbf{User Message Template:}
\begin{lstlisting}
Human Driver Intervention Explanation:
{explanation}

Current Feature Values:
[Same feature list as LM_att with current robot/human values]

Current Reward Weights after a Physical Intervention Update:
- speed_desirability: {current_weight_0:.3f}
- lane_alignment: {current_weight_1:.3f}
- off_road: {current_weight_2:.3f}
- cone_distance: {current_weight_3:.3f}
- car_distance: {current_weight_4:.3f}
- puddle_distance: {current_weight_5:.3f}

Now, for absolutely EVERY feature (considering the explanation, feature changes, and current weights):
1. What absolute change with direction (this will be your 'mu') would support this intervention? Consider the scale of the features, and the current weights.
2. How confident are you in your decision? (confidence score 0.0-1.0)
\end{lstlisting}

\subsection{LLM Configuration Parameters}
\label{app:llm-config}

The table below describes the parameters we used for the LLM experiments using the OpenAI API.

\begin{table}[h]
\centering
\caption{LLM API Configuration Settings}
\begin{tabular}{lcc}
\toprule
\textbf{Parameter} & $\textbf{LM}_{\text{att}}$ & $\textbf{LM}_{\text{pref}}$ \\
\midrule
Model & gpt-4o & gpt-4o \\
Temperature & 0.1 & 0.3 \\
Response Format & JSON & JSON \\
Max Tokens & default & default \\
\bottomrule
\end{tabular}
\end{table}


\begin{table*}[ht]
\centering
\caption{Questionnaire constructs, items, and reliability used in the user study, adapted from Bajcsy et al.~\cite{bajcsy2017phri}.}
\label{tbl:questions}
\begin{tabular}{p{3.5cm} p{9cm} c}
\hline
\textbf{Construct} & \textbf{Items} & \textbf{Cronbach's $\alpha$} \\
\hline
\multirow{2}{*}{Understandability}
  & The robot understood how I wanted to do the task. & \multirow{2}{*}{$\alpha=.953$} \\
  & The robot learned from my corrections. & \\
\hline
\multirow{2}{*}{Ease of Use}
  & The robot required minimal corrections. & \multirow{2}{*}{$\alpha=.816$} \\
  & (Negative) I had to keep correcting the robot. & \\
\hline
\multirow{2}{*}{Predictability}
  & It was easy to anticipate how the robot would respond to my corrections. & \multirow{2}{*}{$\alpha=.750$} \\
  & (Negative) The robot’s response to my corrections surprised me. & \\
\hline
\multirow{2}{*}{Collaborativity}
  & The robot worked with me to complete the task. & \multirow{2}{*}{$\alpha=.928$} \\
  & (Negative) The robot did not collaborate with me to complete the task. & \\
\hline
\end{tabular}
\end{table*}

\section{Additional Simulated Experiment Details}
\label{app:experiments}


\subsection{Feature Definitions and Computations}
\label{app:feature-defs}

Below is a table of formulas used to calculate the features we used to represent a user's preference. Here, $\sigma$ refers to the sigmoid function.

\begin{table}[H]
\centering
\caption{Exact Feature Definitions Used in Experiments}
\begin{tabularx}{\linewidth}{
  >{\raggedright\arraybackslash}p{2cm}
  >{\raggedright\arraybackslash}X
}

\toprule
\textbf{Feature} & \textbf{Mathematical Definition} \\
\midrule
speed\_desirability & $1.0 - \left(\frac{v - v_{\text{target}}}{v_{\text{target}}}\right)^2$ \\
\addlinespace
lane\_alignment & $1-\left(\frac{d_{\text{lane}}}{\text{lane\_width}}\right)^2$ \\
\addlinespace
off\_road & $1.0$ if in bounds, else
$1.0 - \left(\frac{d_{\text{boundary}}}{\mathrm{lane\_width}}\right)^2$ \\
\addlinespace
cone\_distance &
$\min\!\left(1-\frac{r_{\text{safe}}-d_{\text{cone}}}{r_{\text{safe}}}\cdot
\sigma\!\left(-\gamma\cdot|\mathrm{horiz.\_distance}|\right)\right)$ \\
\addlinespace
car\_distance &
$\min\!\left(1-\frac{r_{\text{safe}}-d_{\text{car}}}{r_{\text{safe}}}\cdot
\sigma\!\left(-\gamma\cdot|\mathrm{horiz.\_distance}|\right)\right)$ \\
\addlinespace
puddle\_distance &
$\min\!\left(1-\frac{r_{\text{safe}}-d_{\text{puddle}}}{r_{\text{safe}}}\cdot
\sigma\!\left(-\gamma\cdot|\mathrm{horiz.\_distance}|\right)\right)$ \\
\bottomrule
\end{tabularx}
\end{table}

\subsection{Ground Truth Reward Weights}
\label{app:gt-weights}

The ground-truth reward weights define the ``optimal" behavior of the vehicle that we aim to achieve. The table below shows the values associated with each feature that we used in our experiments.

\begin{table}[H]
\centering
\caption{Ground Truth Preference Weights ($\theta^*$) for Each Scenario}
\resizebox{\linewidth}{!}{%
\begin{tabular}{lcccccc}
\toprule
\textbf{Scenario} & \textbf{Speed} & \textbf{Lane} & \textbf{Off-Road} & \textbf{Cone} & \textbf{Car} & \textbf{Puddle} \\
\midrule
C (Cone) & 5.0 & 2.5 & 20.0 & 40.0 & --- & --- \\
CP (Cone+Puddle) & 5.0 & 1.5 & 10.0 & 20.0 & --- & 1.0 \\
CPC-3 & 5.0 & 2.5 & 20.0 & 40.0 & 50.0 & 3.0 \\
CPC-4 & 5.0 & 2.5 & 20.0 & 40.0 & 50.0 & 3.0 \\
\bottomrule
\end{tabular}
}
\end{table}

\subsection{Hyperparameter Settings}
\label{app:hyperparams}

Our experiments used the following hyperparameters for running the InterACT lab driving simulator.

\begin{table}[H]
\centering
\caption{Complete Hyperparameter Configuration}
\begin{tabular}{lcc}
\toprule
\textbf{Parameter} & \textbf{Symbol} & \textbf{Value} \\
\midrule
Base learning rate & $\alpha$ & 1.0 \\
Language confidence scale & $\sigma$ & 1.2 \\
Numerical stability & $\epsilon$ & $10^{-4}$ \\
Prior stability & $\epsilon_{\text{prior}}$ & $10^{-6}$ \\
Variance stability & $\epsilon_{\text{var}}$ & $10^{-3}$ \\
Capping factor & --- & 5.0 \\
Beta power & $p$ & 2.0 \\
Simulation frequency & --- & 30 Hz \\
Planning horizon & --- & 5 timesteps \\
Episode length & --- & 220 timesteps \\
\bottomrule
\end{tabular}
\end{table}

\subsection{Computational Requirements}
\label{app:compute}

Our simulated results require the following hardware considerations to be satisfied in order to run.

\begin{table}[H]
\centering
\caption{Computational Analysis}
\begin{tabular}{lc}
\toprule
\textbf{Metric} & \textbf{Value} \\
\midrule
Hardware & CPU \\
Per-episode runtime & $\sim$1 minute \\
LLM API calls per intervention & 2 (LM$_{\text{att}}$ + LM$_{\text{pref}}$) \\
Storage per experiment & $<$2MB (logs + trajectories) \\
\bottomrule
\end{tabular}
\end{table}

\subsection{Natural Language Utterances}
\label{app:utterances}

The five natural language utterances used across all experiments, ordered by specificity:

\begin{enumerate}
\item \textbf{"Be careful."} (most ambiguous)
\item \textbf{"Watch out for that thing."} (moderate ambiguity)
\item \textbf{"Stay away from that thing."} (moderate specificity)
\item \textbf{"Avoid the obstacle."} (specific action)
\item \textbf{"Stay away from construction zones."} (specific action)
\item \textbf{"Steer clear of the cone."} (most specific)
\end{enumerate}

\subsection{Intervention Configuration}
\label{app:interventions}

\textbf{Intervention Windows:} Human interventions in our simulated experiments occur during these timestep ranges:
\begin{itemize}
\item Intervention 1: timesteps 45--55
\item Intervention 2: timesteps 85--95
\item Intervention 3: timesteps 130--140
\item Intervention 4: timesteps 170--180
\end{itemize}

\textbf{Correction Generation:} Simulated human trajectory generated by running the planner with the ground truth preference weights.

\subsection{Critical Implementation Details}
\label{app:impl-details}

\textbf{$\mu$ Capping Mechanism:} To prevent instability, LLM-generated preference shifts are capped:
\begin{equation}
\mu_i^{\text{capped}} = \text{sign}(\mu_i) \cdot \min(|\mu_i|, 5.0 \cdot |\Delta\Phi_i|)
\end{equation}

\textbf{Feature Normalization:} Features are normalized such that their highest value is 1 and they decrease to approximately 0 in undesirable situations.

\subsection{Analysis of Physical Context and Ablations}
\label{app:ablation-context}

To further evaluate the necessity of physical context in language interpretation, we performed a comparison between our full model and a stricter ``No-Context'' baseline. In our primary simulated experiments, the ``Language Only'' ablation is context-aware, meaning the LLM receives the feature differences ($\Delta\Phi$) induced by the physical correction to help ground the utterance. 

\para{No-Context Baseline:} During system development, we evaluated a version of the model where the LLM received the language utterance $l$ but no information regarding the physical state changes or current rewards. As shown in Fig.~\ref{fig:no-context-nmse}, without this privileged physical context, the model fails to disambiguate vague commands (e.g., ``Careful!'', ``Stay Away!''). This leads to significantly higher NMSE compared to the full QuickLAP framework, as the model lacks the grounding necessary to map abstract words to specific reward features.

\para{Experimental Scope:} We note that a baseline without any physical corrections at all would fundamentally change the task; in our interactive setting, language is produced specifically to explain or modulate an ongoing physical correction. Therefore, the context-aware ``Language Only'' ablation used in Section V is the most informative baseline, as it isolates the specific benefit of our proposed Bayesian fusion rule (Eq.~\eqref{eq:map-update}).

\para{Simulation vs. Real-World Divergence:} While the context-aware ``Language Only'' variant is competitive in simulation, our user studies (Fig. 5 in the main text) demonstrate that it performs significantly worse with real users. In practice, human language is often high-level (e.g., ``move to the right lane'') and does not map directly onto the robot's pre-defined features. In these cases, QuickLAP’s ability to ground the update in the physical signal while using language to clarify intent provides the robustness necessary for real-world HRI.

\begin{figure}[H]
    \centering
    \includegraphics[width=0.8\linewidth]{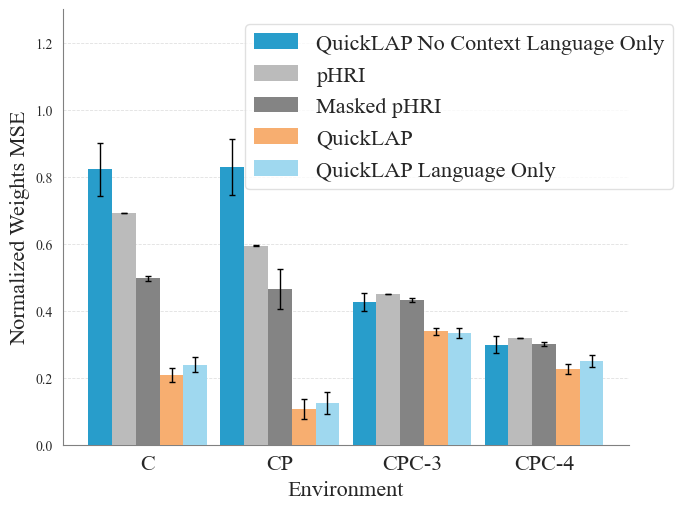}
    \caption{NMSE in the Driving Domain comparing the No-Context baseline against context-aware variants. Without physical grounding, the LLM cannot reliably map vague utterances to the correct reward features.}
    \label{fig:no-context-nmse}
\end{figure}

\begin{figure}[H]
    \centering
    \includegraphics[width=0.8\linewidth]{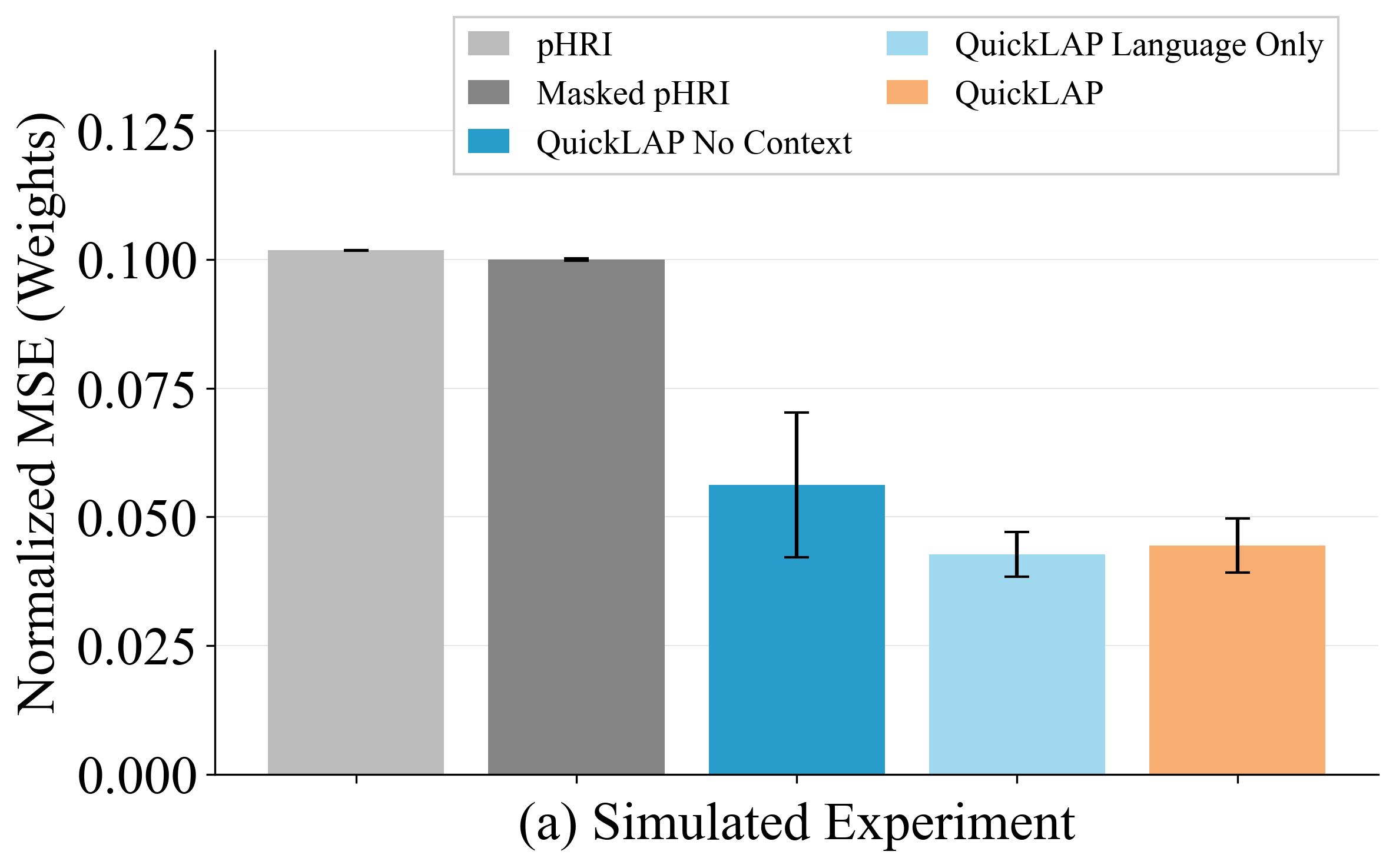}
    \caption{NMSE in the Robot Manipulation Domain comparing the No-Context baseline against context-aware variants. Without physical grounding, the LLM cannot reliably map vague utterances to the correct reward features.}
    \label{fig:no-context-nmse}
\end{figure}
\subsection{Per-Scenario Performance Breakdown}
\label{app:scenario-breakdown}

We reported aggregated NMSE scores in the main text. Here we provide the marginal NMSE values for each environment and each baseline in our simulated experiments.

\begin{table}[H]
\centering
\caption{Normalized MSE by Scenario (mean $\pm$ SEM across 6 utterances for 4 intervention experiment)}
\resizebox{\linewidth}{!}{%
\begin{tabular}{lccc}
\toprule
\textbf{Scenario} & \textbf{Physical-Only (pHRI)} & \textbf{Masked dPhi} & \textbf{QuickLAP} \\
\midrule
C (Cone) & $0.6923 \pm 0.0000$ & $0.4783 \pm 0.0116$ & $\mathbf{0.2071 \pm 0.0453}$ \\
CP (Puddle) & $0.5941 \pm 0.0013$ & $0.3034 \pm 0.0236$ & $\mathbf{0.0761 \pm 0.1090}$ \\
CPC-3 & $0.4496 \pm 0.0000$ & $0.4177 \pm 0.0004$ & $\mathbf{0.3468 \pm 0.0331}$ \\
CPC-4 & $0.3189 \pm 0.0000$ & $0.2876 \pm 0.0003$ & $\mathbf{0.2058 \pm 0.0525}$ \\
\bottomrule
\end{tabular}
}
\end{table}

\subsection{Language Robustness Analysis}
\label{app:language-robustness}

We performed an analysis of different kinds of feedback that a person may provide in simulation to test the robustness of our approach to different ways of communicating with the vehicle. We report our analysis in \autoref{tab:utterance_nmse}.

\begin{table}[t]
\centering
\resizebox{\linewidth}{!}{%
\begin{tabular}{lccc}
\toprule
\textbf{Utterance$^\dagger$} & \textbf{Masked} & \textbf{QuickLAP} & \textbf{pHRI} \\
\midrule
Avoid obstacle & $0.3687$ & $\mathbf{0.1877}$ & $0.5137$ \\
Be careful & $0.3870$ & $\mathbf{0.2772}$ & $0.5137$ \\
Stay away (construction) & $0.3687$ & $\mathbf{0.2087}$ & $0.5137$ \\
Stay away (object) & $0.3687$ & $\mathbf{0.2068}$ & $0.5137$ \\
Steer clear (cone) & $0.3687$ & $\mathbf{0.1672}$ & $0.5137$ \\
Watch out (object) & $0.3687$ & $\mathbf{0.2062}$ & $0.5100$ \\
\bottomrule
\end{tabular}}
\caption{NMSE across methods for different natural language utterances.
$^\dagger$Utterances abbreviated; full forms provided in text.}
\label{tab:utterance_nmse}
\end{table}

\subsection{Additional User Study Details}
\label{appendix:user_study_details}

We present the additional user study details here to enable future research to replicate our findings.

\subsubsection{Manipulation Scenario User Study}
Participants interacted with a Franka Emika Panda robot arm in a robosuite simulation using a SpaceMouse to provide corrections to the robot, while verbally describing why they provided their correction. The interaction was run using the CPU of a single laptop. Users' verbal explanations were captured using the laptop's built-in microphone, and transcribed using the OpenAI Whisper speech recognition model.  

\subsubsection{Car Scenario User Study}
Participants controlled the robot using the Logitech steering wheel controller with three pedals. Participants were instructed to use the wheel as well as the gas pedal and brake pedal to control the vehicles behavior. The simulated environment was run using the CPU of a single laptop. Participants' voice commands were recorded using the laptop's built-in microphone, and were transcribed using the OpenAI Whisper speech recognition model.

\textbf{Questionnaire.}
Our user study adapted the questionnaire from Bajcsy et al.~\cite{bajcsy2017phri} with small modifications. The exact Likert items used in the study are described in Tbl.~\ref{tbl:questions}. Each question was rated on a seven-point scale from strongly disagree to strongly agree. We additionally provide the interrater reliability scores for each construct as found in our study.

In addition to the Likert questions, we asked users to rank all three algorithms that they experienced in a forced-choice ranking question. Participants were not allowed to specify that two algorithms tied. This question was phrased as ``Please use the algorithm codes (1,2,3) to rank the algorithms'' and participants were provided with three boxes arranged vertically to place the numbers one through three. The top box was labeled ``Most Preferred'' and the bottom box was labeled ``Least Preferred''.

Participants were also provided a space to take notes on each of the three algorithms. These results were not analyzed in this paper, but allowed participants to keep track of the impressions of the three algorithms throughout the study.


\subsection{Robosuite Pick-and-Place Experiment Details}
\label{app:robosuite-details}

We ran simulated learning experiments in a robosuite pick-and-place environment to evaluate QuickLAP for manipulation.

\para{Task.}
The robot picks a red block from Zone A and places it into Zone B. A green block and Zone C act as obstacles. Following standard robosuite pipelines, we decompose the behavior into phases (approach, grasp, lift, transport, place, release); only the \emph{transport} phase uses MPC optimization with learned reward weights, while the other phases are executed with direct (non-optimized) controllers.

\para{Reward features.}
The reward is a weighted sum of seven features: green clearance, velocity, collision safety, joint safety, block-to-target zone proximity, Zone-C clearance, and height maintenance (Table~\ref{tab:robosuite-features}).

\begin{table}[H]
\centering
\caption{Robosuite feature names and short descriptions.}
\begin{tabularx}{\linewidth}{lX}
\toprule
\textbf{Feature} & \textbf{Description} \\
\midrule
green\_clearance & Clearance from the green block; higher = farther. \\
velocity & End-effector speed; higher = faster. \\
collision\_safety & Collision avoidance; higher = more cautious. \\
joint\_safety & Joint limit margin; higher = safer. \\
block\_to\_target\_zone & Proximity of the block to target Zone B; higher = closer. \\
zone\_c\_clearance & Clearance from obstacle Zone C; higher = farther. \\
height\_maintain & Adherence to a nominal transport height; higher = closer. \\
\bottomrule
\end{tabularx}
\label{tab:robosuite-features}
\end{table}

\para{Ground-truth weights.}
We define oracle (expert) weights $\theta^*$ for evaluation as:
green clearance 10.0, velocity 1.0, collision safety 2.0, joint safety 1.0, block-to-target zone 25.0, Zone-C clearance 2.0, height maintenance 0.0.

\para{Utterances.}
We use six natural language utterances spanning varying specificity:
``Avoid the obstacle.'', ``Steer clear of there.'', ``Move away!!'', ``AHH!'', ``Go left'', ``Get away''.

\para{Baselines and logging.}
We compare (i) \textbf{naive} (physical-only pHRI update), (ii) \textbf{masked\_dphi} (LLM-masked $\Delta \Phi$), and (iii) \textbf{adapt\_gated\_llm} (QuickLAP: physical + language using attention and confidence).
Experiments are run with \texttt{run\_experiments.py}; metrics (normalized weight MSE and regret) are logged to \texttt{logs/}.

\end{document}